\pdfoutput=1
\documentclass{article}

\usepackage[preprint]{tmlr}

\usepackage{microtype}

\usepackage{amsmath}
\usepackage{amssymb}

\usepackage{booktabs}
\usepackage{float}
\usepackage{placeins}

\usepackage{graphicx}
\usepackage[labelfont=bf]{caption}
\usepackage{subcaption}
\usepackage{tikz}
\usetikzlibrary{arrows.meta,positioning,shapes.geometric}
\usepackage{pgfplots}
\pgfplotsset{compat=1.18}
\definecolor{curveblue}{HTML}{2C6FB5}

\usepackage{hyperref}
\hypersetup{
  colorlinks=true,
  linkcolor=black,
  citecolor=blue!60!black,
  urlcolor=blue!60!black,
  pdftitle={PureTD: Reinforcement Learning for Backgammon Money Games with No Evaluation-time Search},
  pdfauthor={Alexander Strehl},
}

\usepackage{algorithm}
\usepackage{algorithmic}
\newcommand{\COMMENTLINE}[1]{\STATE // \textit{#1}}

\usepackage{xspace}
\usepackage{xcolor}

\newcommand{\gnubg}{\texttt{gnubg}\xspace}

\title{PureTD: Reinforcement Learning for\\
  Backgammon Money Games with No Evaluation-time Search}

\author{%
\name Alexander Strehl \email astrehl@gmail.com \\
\addr Independent Researcher
}

\begin{document}
\maketitle

%%%%%%%%%%%%%%%%%%%%%%%%%%%%%%%%%%%%%%%%%%%%%%%%%%%%%%%%%%%%%%%%%%%%%%%%%%%%%%%%
\begin{abstract}
We revisit Tesauro's TD-Gammon for backgammon money games in the setting of no
evaluation-time search.  Both
checker play and cube action (use of the doubling cube) are learned from scratch via self-play
reinforcement learning (RL), with minimal hand-coded logic and no expert features.  In this
setting, we demonstrate that pure self-play RL suffices to train models that
reach near-state-of-the-art playing strength.  Specifically, for cubeful
money games, our search-free model evaluates faster and is substantially
stronger than the open-source engines GNU Backgammon and Open Sage
running a one-move (1-ply) look-ahead search.  Our method of training a
cubeful model also outperforms the standard approach of training a cubeless model and deriving cubeful equity
with Janowski-style formulas, when the two models are equally sized.
\end{abstract}

%%%%%%%%%%%%%%%%%%%%%%%%%%%%%%%%%%%%%%%%%%%%%%%%%%%%%%%%%%%%%%%%%%%%%%%%%%%%%%%%
\section{Introduction}
\label{sec:intro}
Computer backgammon engines combine two components at decision time:
a learned \emph{position evaluator}, typically a neural network that maps a
board state to an equity (we use \emph{evaluator}, \emph{model}, and
\emph{network} interchangeably), and an \emph{expectimax search} that computes an
expectation of the evaluator's scores over the game states reached after $n$
further player turns.  In the shallowest possible search, for a given board state, an
action is selected by querying the evaluator on the resulting position for each
candidate move and picking the move that maximizes the evaluator's score. Throughout this paper we use the
terminology in which ``0-ply'' describes this procedure, where the evaluator is
consulted directly with no look-ahead, and ``$n$-ply'' means a look-ahead
expanding $n$ player turns. We
focus primarily on the 0-ply setting.  Concretely, we ask: how
strong can a backgammon agent be made if its decisions are produced by a
single forward pass of a neural network per candidate move?

While the strength of an evaluator is worth studying in isolation, maximizing
it is not the objective for a backgammon engine, which must optimize
the full system (evaluator combined with search).  Viable engines such as GNU Backgammon (\gnubg)~\citep{gnubg}
must therefore balance stronger but larger networks against
smaller ones that evaluate faster during search.  This paper, by
contrast, optimizes a single evaluator's strength.
However, our models also offer a favorable trade-off between playing strength
and computational cost: for
both cubeless and cubeful money games, our 0-ply models are stronger than, and
at least $3.5\times$ faster than, the open-source engines \gnubg{} and Open Sage~\citep{bgsage} running a 1-ply
search.\looseness=-1

Despite its limitations, a study of the 0-ply setting remains valuable for a
couple of reasons.  First, a stronger evaluator is broadly expected to yield a
stronger engine once combined with search, and capping network size prematurely
risks limiting what the full system can ultimately achieve.  Second, the 0-ply setting is of interest in its own right, in the spirit of
searchless chess~\citep{ruoss2024amortized}, and offers a useful data point on the
ceiling reachable by self-play RL.

\paragraph{Money games and the doubling cube.}
Backgammon money games and match play include the \emph{doubling cube}, which
starts at a value of $1$.  Initially the cube is ``centered'' and owned by
neither player.  At the beginning of a turn and before the dice
are rolled, if the cube is centered or owned by the player on roll, they can
\texttt{offer\_double}, proposing to double the cube value and with it the
stakes.  The opponent must either accept (\texttt{take}), continuing the game
at the doubled value and taking ownership of the cube, or refuse
(\texttt{pass}), conceding the current cube value and ending the game.
A money game ends either when a double is passed or when a player bears off
all of their checkers.  In the latter case, the game is scored as a normal win,
a gammon, or a backgammon, in which the winner gains $c$, $2c$, or $3c$
respectively and the loser loses the same amount, where $c$ is the current cube
value.  A passed double is worth $c$.  In cubeless
play, $c = 1$ throughout.  The \emph{equity} of a position is a player's
expected net winnings from it.\footnote{With an unlimited doubling cube, this
expectation may not exist for every position.  Positions have been
constructed in which repeated redoubling appears to be correct, which would
leave the equity
undefined~\citep{floyd1982tiger,zare2004undefined,chow2026equity}, although
this conclusion has not been formally proved.}  How a player chooses to handle the cube in a
given position is called the \emph{cube action}.  Finding optimal cube actions
is difficult, since they depend on win and gammon rates, on the cube's current ownership,
and on the recube vigorish, the future value of being able to
redouble~\citep{bkgmrecubevig}.  The inclusion of the doubling cube also changes the optimal checker play
in significant ways~\citep[Ch.~13]{schiemann2026theory}, and the required
adjustments challenge even the strongest players.  For a specific example, see
Appendix~\ref{app:cube-checker}, where Zdenek Zizka, one of the world's top
players, chose a play that dominates the optimal cubeful play on all
six
cubeless components (wins and losses, gammons won and lost, backgammons won and
lost), but is nevertheless a major error.

\paragraph{Open-source pipeline and large-scale evaluation.}
We call our method, and the resulting models, PureTD.  We release a complete training pipeline and evaluate the resulting models in
two complementary ways.  The first is an
offline error-rate metric for cubeful money, eXtreme Gammon's XG++ Performance
Rating (PR, lower is better).  The second is head-to-head play against \gnubg{}
and Open Sage at 0-ply and 1-ply, in both cubeless and cubeful money.

\paragraph{Code and reproducibility.}
Code, trained models, and commands to reproduce the released models and the main
evaluations are released under the MIT license at
\url{https://github.com/alexstrehl/puretd}.
Results correspond to commit \texttt{9c9e5138364a}.
Training code and
preliminary results have been public since April 2026.

\subsection{Contributions}
\label{sec:contributions}

Backgammon is a popular stochastic game with a large branching factor, given
that every turn passes through a chance node with 21 distinct outcomes.
It remains
a natural testbed for RL methods, and discoveries made on it may apply to
similar games in the future.  Recently, \citet{boige2026avi} found that
approximate value iteration in self-play, without tree search, is competitive
with AlphaZero-style training on several deterministic board games.  Our
training with exact-Bellman backups (Sec.~\ref{sec:architecture}) is a form
of approximate value iteration, and our results lend support to their
finding in a stochastic game.  Modern backgammon also includes the doubling cube,
which increases the challenge and complexity.  As one indication of the added
complexity, measured on a game tree in which each edge is one complete turn of the
player on roll, the
branching factor nearly doubles, from approximately 400 for cubeless play to 750 for
cubeful play (Appendix~\ref{app:branching}).

Virtually all of the strongest modern engines, including eXtreme Gammon
(XG)~\citep{xg} and \gnubg{}, use an evaluator
trained on cubeless money games and adapt it to cubeful money games (and match
play) by computing cubeful equity heuristically from Janowski's
formulas~\citep{janowski1993cube, madsen2026janowski}
(Appendix~\ref{app:janowski}).  The main contribution of this paper is to demonstrate that for
money games, a simple RL method suffices to train competitive evaluators that
learn cubeful play end-to-end.

To achieve a cubeful evaluator, we follow \citet{lin2020cube} and extend the
agent's action space with the doubling decision
(\texttt{offer\_double} versus \texttt{no\_double}) at the start of a turn,
treating it as an ordinary state transition learned via self-play RL.
Unlike Lin, we generate a TD update for every cube decision,
including \texttt{no\_double}, and add an input flag that lets the network
distinguish a cube decision from the subsequent checker play
(Sec.~\ref{sec:cube}).  In Sec.~\ref{sec:cube-ablation} we find that this approach
outperforms the standard Janowski-style cube handling on similarly sized
models.

While there has been significant prior work on RL for
backgammon~\citep{tesauro1995tdgammon, antonoglou2022planning}, studies that
measure computational cost against evaluator strength have been limited.  As a
final engine will always combine the evaluator with a run-time search, the
computational cost is critical.  We report playing strength against evaluation
cost for our models and two baselines (Sec.~\ref{sec:cost}).

BGBlitz~\citep{bgblitz} and the author of Octopus (Sec.~\ref{sec:background})
report that self-play RL alone suffices for extremely strong play, but to our
knowledge ours is the first work to publish exact recipes and code
(Sec.~\ref{sec:method}).  We show that with larger networks and a few
modern changes to the learning procedure, specifically ReLU instead of sigmoid
activations, batch training, and exact-Bellman backups, the TD-Gammon procedure
alone is sufficient to produce a near-state-of-the-art 0-ply evaluation function for money
games.  This is not the route many recent
engines have taken.  In 2026 alone, at least four new strong engines appeared
(Sec.~\ref{sec:background}).  The three with documented training, Open Sage~\citep{bgsage},
Raccoon~\citep{madsen2026raccoon}, and Whittington's
Backgammon-NN~\citep{whittington2026backgammonnn}, all use a mature engine,
\gnubg{}, to compute the targets.  All three report trying self-play
RL first and reaching a plateau significantly below their final models' strength.

Finally, there is growing interest in applying more expensive methods to
backgammon, including larger models and more complex and expensive self-play in
the style of AlphaZero~\citep{antonoglou2022planning, madsen2026raccoon}.  Our
results give a baseline for what a relatively simple learning strategy can
achieve, and include preliminary findings on the effect of model size on
evaluator strength (Sec.~\ref{sec:network-size}).

%%%%%%%%%%%%%%%%%%%%%%%%%%%%%%%%%%%%%%%%%%%%%%%%%%%%%%%%%%%%%%%%%%%%%%%%%%%%%%%%
\section{Background and Related Work}
\label{sec:background}

\paragraph{Temporal-difference (TD) learning and the TD-Gammon formulation.}
The standard approach to computer backgammon is to train a model, typically a neural network of
limited depth, that given a position $s$ outputs the approximate value $V(s)$ of
the position for a given player.  At play time, games proceed by picking the
action that the model predicts is best from the current state.  Formally, the
game is a two-player zero-sum alternating Markov
game~\citep{condon1992complexity,littman1996generalized}.  Despite the name,
the players need not strictly alternate.  For example, when the player on roll
can double but chooses not to (Sec.~\ref{sec:cube}), that same player stays on
roll and acts again.  Optimal play then satisfies a Bellman equation of the
same form as for a Markov decision process
(MDP)~\citep[\S3.6]{sutton2018reinforcement}, with $\max$ at one player's
states and $\min$ at the other's.

For ease of exposition, equations in this paper use the convention that
$V(s)$ is the value of state $s$ for a fixed reference player, the
\emph{maximizer}, whose goal is to maximize $V$ at every state, while the
opponent correspondingly minimizes $V$ (equivalently, maximizes $-V$).  Because
the reference player is fixed, no state transition induces a sign change in $V$,
whether or not the on-roll player changes.  This lets every displayed value
equation use a single $\max$, stated for a state with the maximizer on roll.

For an optimal value function $V^*$, Bellman's equation requires
\begin{equation}
\label{eq:bellman}
V^*(s) \;=\; \max_a \; \mathbb{E}_{s' \sim T(s,a)} \bigl[\, V^*(s') \,\bigr],
\end{equation}
where $T(s,a)$ is the transition distribution over next states induced by action $a$.
When the opponent is on roll, the same equation holds with $\min_a$ in place of
$\max_a$, which completes the recursion.\footnote{For an action that ends the current player's turn, expanding one further ply makes the
suppressed minimization explicit: $V^*(s) = \max_a \mathbb{E}_{s' \sim T(s,a)}\!\bigl[\min_{a'}
\mathbb{E}_{s'' \sim T(s',a')}[\, V^*(s'') \,]\bigr]$, where $s'$ has the opponent on roll
(hence the $\min$).  An equivalent way to express all cases in
Eq.~\eqref{eq:bellman} is $V^*(s) = \sigma(s) \max_a \mathbb{E}_{s' \sim
T(s,a)}\bigl[\sigma(s)\, V^*(s')\bigr]$, where $\sigma(s) = +1$ if the maximizer
is on roll at $s$ and $-1$ otherwise.}
The game is undiscounted and has no intermediate rewards.  At a terminal state,
$V^*$ equals the signed outcome for the maximizer (Sec.~\ref{sec:intro}), so
Eq.~\eqref{eq:bellman} has no reward term.

During self-play, the maximizer chooses actions greedily using the current value
estimate $V$,
\begin{equation}
\label{eq:greedy}
a^\star \;=\; \arg\max_a \; \mathbb{E}_{s' \sim T(s, a)}\bigl[\, V(s') \,\bigr],
\end{equation}
and the opponent symmetrically takes $\arg\min$.  The model is trained
incrementally to be consistent with Bellman's equation.  For backgammon
specifically, we found little benefit to any explicit exploration during
training, as natural variation is provided by the dice rolls and by the many
stages of training.

Tesauro's TD-Gammon (\citealp{tesauro1995tdgammon}, \citealp[\S\S6.8, 16.1]{sutton2018reinforcement}) used
a clever technique to reduce the variance of the TD training targets, based on
decomposing a state into a stochastic dice roll and a deterministic
part.  Define a \emph{pseudo-state} $x$ as a backgammon position
\emph{before} the dice are rolled, closely related to the afterstates
of~\citet[\S6.8]{sutton2018reinforcement}.  Rolling $d$ from $x$ yields a state
$s_{x,d}$, from which the player on roll selects a checker play $a$.  For
any pseudo-state $x$,
\begin{equation}
\label{eq:pseudostate}
V^*(x)
\;=\; \mathbb{E}_{d \sim \mathcal{D}}\bigl[\, V^*(s_{x,d}) \,\bigr]
\;=\; \mathbb{E}_{d \sim \mathcal{D}}\bigl[\, \textstyle\max_a Q^*(s_{x,d}, a) \,\bigr]
\;=\; \mathbb{E}_{d \sim \mathcal{D}}\bigl[\, \textstyle\max_a V^*\!\bigl(\tilde{T}(s_{x,d}, a)\bigr) \,\bigr],
\end{equation}
where $\mathcal{D}$ is the distribution over the 21 distinct dice rolls,
$\tilde{T}(s, a)$ is the deterministic next pseudo-state reached by
playing action $a$ in state $s$, and $Q^*(s, a) = V^*\!\bigl(\tilde{T}(s, a)\bigr)$
is the value of playing $a$ in state $s$, which equals the expectation
$\mathbb{E}_{s' \sim T(s,a)}[V^*(s')]$ of Eq.~\eqref{eq:bellman}.  Evaluators are trained directly on
pseudo-states, which cleanly reduces input dimensionality and the
complexity of learning by avoiding any encoding of the dice.

The decomposition of Eq.~\eqref{eq:pseudostate} suggests two ways to form a regression target for $V(x)$: a
\emph{sampled-Bellman target} that samples a single dice roll $d \sim
\mathcal{D}$ and backs up the value of the greedy continuation, and an
\emph{exact-Bellman target} that instead averages over all 21 rolls (the
right-hand side of Eq.~\eqref{eq:pseudostate}), removing the dice variance at the
cost of enumerating every roll.  We define both targets formally in
Sec.~\ref{sec:architecture} (Eqs.~\eqref{eq:sampled} and~\eqref{eq:exact-target}).

\paragraph{Cube normalization.}
In practice, for a fixed state $s$, our model produces a value
$V^{\text{model}}(s)$ representing the equity of the position for the player on
roll per unit of the current cube value $c(s)$.  Thus there is a direct
correspondence between the model and $V$ as used in the equations:
$V(s) = c(s)\,V^{\text{model}}(s)$ when the maximizer is on roll and
$V(s) = -c(s)\,V^{\text{model}}(s)$ otherwise.  This normalization keeps outputs in a
stable numeric range across cube levels and encodes the theoretical fact that
money-game equity scales linearly with the cube, and thus the optimal checker
play and cube decisions are identical at every cube value of $2$ or
greater~\citep{xgmanual, lin2020cube}.

\paragraph{TD-Gammon's historical ceiling.}
The canonical pure-self-play RL backgammon agent is Tesauro's
TD-Gammon~\citep{tesauro1992practical, tesauro1995tdgammon}, which trained a
shallow feed-forward network with TD$(\lambda)$~\citep{sutton1988learning}.  It was the strongest
backgammon program of its time and a major breakthrough.
Subsequent engines
surpassed TD-Gammon's level, but typically only by replacing self-play TD with
more specialized training (see \emph{RL vs. SL training for backgammon} below).
TD-Gammon included a $\lambda$ parameter that controlled the time horizon used
for training targets.  TD-Gammon with $\lambda = 0$ is essentially the
sampled-Bellman variant of the above method where we train the model after every action during
self-play via online stochastic gradient descent.  We did not experiment with
non-zero $\lambda$ values.  Tesauro himself later found $\lambda = 0$
competitive with small non-zero values at roughly half the computational cost
per step, and used it for most subsequent TD-Gammon
training~\citep{tesauro2002programming}.

\paragraph{GNU Backgammon.}
\gnubg{}~\citep{gnubg} is an open-source backgammon engine
that is a standard baseline.  Its evaluator consists of three specialized
single-hidden-layer neural networks ($\sim$30k weights each) for
distinct game phases (contact, race, crashed) with hand-engineered
features.\footnote{A contact position is one in which future hits are possible.
A race is a non-contact position.  A crashed position is a technical subtype of
contact position.}  The engine also
uses a precomputed bearoff database for common endgame positions.  At run
time its networks are combined with $n$-ply expectimax search.

\paragraph{Open Sage.}
Open Sage (\texttt{bgsage})~\citep{bgsage} is a recently released
and actively developed open-source engine.  Its evaluator is an ensemble of
21 distinct neural networks specialized to different
game phases.  Each has a single hidden layer, with roughly 100k parameters for
the more common contact positions and 20k for pure races.  Its documented
training pipeline relies on \gnubg{}'s published
rollout databases, so unlike a pure self-play system, its training
methodology depends on an existing strong engine.  Like \gnubg{}, it relies on
multi-ply search and computes cube decisions from a cubeless evaluator rather
than learning them.  Its developer reports rough parity with XG,
measured at 2-ply truncated-rollout
evaluation (Open Sage's ``3T'' setting in its own ply
convention)~\citep{bgsage2026xgcomparison}.
We caution, however, that strength
measured at a fixed search depth or rollout budget does not by itself establish
a stronger engine, since it omits the per-decision computational cost.

\paragraph{RL vs. SL training for backgammon.}
Neither \gnubg{} nor Open Sage is trained by self-play TD alone. The technique,
documented for both Open Sage~\citep{bgsage} and \gnubg{}~\citep{hauk2006minimax}, is to first
train an evaluator with TD self-play, then refine it
on a selected set of positions whose targets are computed by averaging repeated rollouts
of an existing model, building on the rollout-based policy improvement of
\citet{tesauro1996online}.
This refinement is itself a form of RL: a Monte Carlo
value estimate whose variance is reduced by averaging rollouts, though
it is often referred to as a supervised learning (SL) phase. Restricting refinement to a selected
set of positions is reminiscent of the emphasis on hard examples
in boosting-style training.
PureTD is trained entirely by self-play TD learning, using sampled- and
exact-Bellman targets (Sec.~\ref{sec:architecture}).

\paragraph{The doubling cube and Janowski's formulas.}
Janowski's analytic formulas~\citep{janowski1993cube, madsen2026janowski}
compute approximate cubeful equity and double/take points from cubeless
win equity and gammon rates.  They interpolate between a dead-cube model and
the continuous (live-cube) model of \citet{keeler1975doubling} via an
empirically tuned cube-life index $x$.  Most engines, including \gnubg{} and
XG~\citep{xgmanual}, learn evaluators that model cubeless equity and apply heuristics based on these formulas at inference time, treating cube decisions as a
post-hoc adjustment rather than something the network itself learns.
Specifically, \gnubg{} computes cubeful equity as a piecewise-linear
function of the cubeless win probability, with breakpoints determined by
Janowski's formulas (Appendix~\ref{app:janowski}).
Note that applying these heuristics to a cubeless money model yields an
approximation of the cubeful equity for any position, and thus provides a
method for determining both cube action and cube-aware checker play from a
cubeless model.

\paragraph{Prior RL-native cube work.}
The closest prior work is \citet{lin2020cube}, who first proposed integrating
cube actions into the agent's action space and reported results for match play.
Our work differs in several ways.  First, we fix a key conceptual error in
the formulation that leads to suboptimal play (Sec.~\ref{sec:cube}).  Second,
Lin's results are not measured against a Janowski-style baseline or a strong
engine such as \gnubg.  Third, Lin used networks with a single hidden layer of
roughly 48k weights.  Our 563k-parameter cubeful model is an order of magnitude
larger, trained with modern techniques including exact-Bellman backups, and
evaluated directly on cubeful money play against both \gnubg and a
Janowski-style baseline.

\paragraph{Exact-Bellman updates.}
\citet{hilton2019backgammon} experimented with exact-Bellman backups.  He argued that
the technique is naturally situated
between TD-Gammon's sampled targets and the much slower but more accurate targets used by AlphaZero's MCTS~\citep{silver2018alphazero}.  We use this
technique as a refinement phase in our training pipeline
(Sec.~\ref{sec:architecture}).

\paragraph{Earlier self-play RL studies.}
Tesauro's TD-Gammon~\citep{tesauro1992practical,tesauro1995tdgammon}
demonstrated the effectiveness of self-play RL in backgammon, and subsequent
studies explored variations in architecture and training.
\citet{boyan1992modular} found that modular networks selected by hand-coded
position categories outperformed monolithic networks.
\citet{wiering2010selfplay} found that self-play TD learned faster than passive
observation of expert games.
\citet{papahristou2011variants,papahristou2012improving} applied self-play TD
learning to the backgammon variants Plakoto and Fevga and investigated
improvements to the training procedure.

\paragraph{Other modern backgammon RL.}
XG is closed-source and widely regarded as one of the
strongest available backgammon engines.  The details of its training are not
publicly documented.
BGBlitz~\citep{bgblitz}, a closed-source engine whose performance is similar to
\gnubg{}, and Octopus (Ø.~Schønning-Johansen, personal communication, 2026), an
unreleased model, both report reaching super-human strength through
\emph{self-play RL training}.  Both train a
cubeless model and extend to cubeful play with Janowski's formulas rather than
learning cube decisions directly (personal communications, 2026).

Stochastic MuZero~\citep{antonoglou2022planning} extends MuZero's
learned-model MCTS planning to stochastic environments.  In backgammon it
matches an AlphaZero baseline that plans with a true simulator, and exceeds
\gnubg{} 3-ply with large searches, though it stops short of a full evaluation
that accounts for run-time cost.  It does not handle cubeful money
play, however, and its implementation is not publicly available.

Two other open-source projects are noteworthy:
\texttt{wildbg}~\citep{wenderdel2024wildbg} (supervised plus
search) and \citet{hilton2019backgammon} (RL, discussed above).
\texttt{wildbg}'s training-pipeline design informed our model architecture.

\paragraph{Recent engines.}
Several extremely strong backgammon engines were released during the preparation of our work
and are still under development.  To the best of our knowledge, none of them uses a
learned cubeful model.  The published models are cubeless, and cubeful play
in the released code is derived from Janowski-style formulas.

HedgeHog~\citep{hedgehog2026} is self-described as ``an NNUE-style neural-network evaluator with incremental
accumulation and Highway SIMD.''  Its developer maintains a public
repository with open-source code (MIT license) for evaluating HedgeHog
models.  The full engine is not open source.  The repository was first
published on May 21, 2026, and gained support for the Jacoby rule on
August 10, 2026.
HedgeHog's published
benchmarks~\citep{hedgehog2026benchmarks} report that, at 0-ply over $10^7$
cubeful money games against \gnubg{}'s network running inside HedgeHog's
engine, Fox (version 0.34) wins $31.5$ milli-equity per game (mEq/game) and the
premium Colossus 2.0 model wins $70.8$ mEq/game.  No
training code or methodology is publicly available for HedgeHog.  As a
newer engine it is not one of our primary baselines, but
Appendix~\ref{app:hedgehog} reports strength and evaluation
cost of its two strongest publicly available models under our benchmarks.

Raccoon~\citep{madsen2026raccoon} is an ongoing research project and
backgammon engine built around a 10-block, 256-channel residual network.  Its
developer reports that pure self-play from scratch plateaued well short of
\gnubg{}, and that its published results instead come from distillation of
\gnubg{} 2-ply~\citep{madsen2026raccoon}.  The resulting network shows very
strong 0-ply performance
at significant run-time cost, winning about 47 mEq/game against \gnubg{}
0-ply over $18{,}000$ cubeless money games~\citep{madsen2026raccoonresearch} while
evaluating roughly $3{,}500$ times fewer positions per second than
\gnubg{}~\citep{madsen2026raccoonspeed}.  Cubeful play, without the Jacoby
rule, was added to the repository in September 2026.

Whittington's Backgammon-NN~\citep{whittington2026backgammonnn}, released in
July 2026, trains its current network on $22.5$M positions labeled by
\gnubg{}'s 2-ply evaluation, after earlier self-play networks plateaued.  The
author reports a loss of 113 mEq/game at 0-ply against \gnubg{} 2-ply.  It is distributed as a free
binary, with source available for viewing but not under an open-source license.

Rather than benchmark every available engine, we compare against the strongest
open-source engines supporting cubeful money-game play with the Jacoby rule as of
July 15, 2026, our cutoff for selecting baselines.  To the best of our
knowledge, these were \gnubg{} and Open Sage.  For both we use the latest
release available when we ran our experiments (Appendix~\ref{app:repro}).

%%%%%%%%%%%%%%%%%%%%%%%%%%%%%%%%%%%%%%%%%%%%%%%%%%%%%%%%%%%%%%%%%%%%%%%%%%%%%%%%
\section{Method}
\label{sec:method}

\subsection{Board Encoding}
\label{sec:encoding}

For the feature representation, we adopt Tesauro's per-point thermometer board encoding
(\citealp{tesauro1992practical}, as described in \citealp[\S11.1]{sutton1998reinforcement}):\footnote{Tesauro's
own description is brief: ``a truncated unary encoding with four units was
used.  The first three units encoded separately the cases of one man, two men,
and three men, while the fourth unit encoded the number of men beyond 3''
\citep[Sec.~3.4, p.~270]{tesauro1992practical}.  We could not verify the exact
encoding from his publications, including the scaling of the fourth unit and
of the bar and borne-off units.  We use the cumulative form described in the
first edition of \citet[\S11.1]{sutton1998reinforcement}.}
four features per point per player, namely binary units for one-or-more,
two-or-more, and three-or-more checkers and a fourth feature encoding the
number of additional checkers (the surplus beyond three, halved).  There are
also two scalar features per player, the number of checkers on the bar
divided by $2$ and the number borne off divided by $15$, for a total of 196
base features.
We use a \emph{perspective encoding}, always calling the network from the
perspective of the side to play and moving in a fixed direction.  Although a minor
detail, this choice encodes a basic symmetry of backgammon.  For
example, every backgammon
player knows to ``make the 5-point'' on a first roll of $(3,1)$ regardless
of whether they are moving clockwise or counter-clockwise.  The cubeful model
has four additional cube-specific features as described in
Sec.~\ref{sec:cube}.

\subsection{Architecture and Training Pipeline}
\label{sec:architecture}

\paragraph{Architecture.}
Our models are multilayer perceptrons (MLPs) implemented in PyTorch.  The cubeful money model
has 200 inputs feeding four hidden layers of sizes $[512, 512, 256, 256]$, for
a total of 563k parameters.  All hidden layers use ReLU activations, replacing the
sigmoid units of the original TD-Gammon.  As expected, we found that
increasing the network size improves playing strength at a higher
evaluation cost.  The current topology was chosen as a subjective balance
between the two, and Sec.~\ref{sec:network-size}
reports preliminary experiments with other sizes.

\paragraph{Output heads.}
For cubeful money, the output is linear and
interpreted as \emph{equity per unit cube value}, and realized equity is
\texttt{output} $\times$ \texttt{current\_cube\_value} (see ``Cube normalization'' in Sec.~\ref{sec:background}).

\paragraph{Training loop.}
TD-Gammon trained via online learning, interleaving self-play moves with
stochastic-gradient-descent updates.  On modern hardware we found
it more practical to decouple self-play data collection from model training,
yielding a batch-style pipeline. At each round we play $N$ games ($N = 1{,}000$ in all our reported results) with
the current network as both players, compute a regression target for every
visited state, then run one epoch of minibatch SGD with Adam~\citep{kingma2015adam}.  Rounds repeat
with fresh self-play data.  For reference, on a
single high-core-count AMD Zen~5 workstation with a consumer GPU,
training runs at roughly $3{,}800$--$4{,}000$ games/s with
sampled-Bellman backups and $750$--$880$ games/s with exact-Bellman backups.  In total, training a cubeful model of production strength from scratch (the
four-layer $[512, 512, 256, 256]$ network) with the recipe of
Appendix~\ref{app:retrain} requires approximately $250$M training
episodes (${\sim}60$M sampled-Bellman, ${\sim}190$M exact-Bellman), or about $70$
wall-clock hours on such a machine.

\paragraph{Progressive expansion and cross-task transfer.}
The final $[512, 512, 256, 256]$ network was trained incrementally, reusing
weights across both architectures and tasks.  Beginning from a single-hidden-layer network, we alternately
\emph{widened} and \emph{deepened} the network in stages
($[80] \to [150] \to [150, 150] \to [256, 256] \to [256, 256, 256] \to
[512, 512, 256] \to [512, 512, 256, 256]$), each time transferring the existing
parameters into the larger network.  Widening adds units whose connections into the
existing units and the output have zero-initialized weights, which preserves
the function exactly, and deepening inserts an
identity-initialized layer in the style of Net2Net~\citep{chen2016net2net},
which preserves the function up to a small symmetry-breaking perturbation.  We
also transferred across tasks: a double-match-point (DMP, i.e.\ one-point match) network warm-started a
four-layer cubeless base, reusing the hidden layers while re-initializing
the output head to predict equity, which in turn warm-started the cubeful-money
model by extending the input layer with the cube features.  A brief
learning-rate warmup after each task transition improved stability: rather
than resuming at the full learning rate, we linearly ramped it from roughly
$10\%$ of its target value up to the target over the first $20$ training
rounds.  All parameters remained trainable at every stage (no weights were
frozen).  We found that this simple
form of transfer reused learned representations and accelerated
training relative to learning each task from scratch.

The production models
were trained over a period of months in many successive stages.  In Appendix~\ref{app:retrain} we give exact settings to train both the
cubeful-money and cubeless (prob5, Sec.~\ref{sec:results}) models from scratch, and
Figure~\ref{fig:retrain-curves} tracks their playing strength over the course
of training.  The cubeful recipe totals about $250$M episodes and the
cubeless recipe about $120$M, and both include a shared base of about $50$M
episodes.  We ran each recipe
twice, and all four resulting models are within about $1$ mEq/game of our
production models (Appendix~\ref{app:retrain}).

Although the production models' performance initially appeared to have
plateaued, further training produced additional gains.  We continued training
the production models for another $500$M episodes, after which they
outperformed their original versions by about $3$--$7$~mEq/game.  Details are presented in Appendix~\ref{app:convergence}, along with a formal
stopping rule.
After the additional training,
learned cube handling kept its advantage over both Janowski-style heuristics
and the Lin-style ablation, with the alternatives trained at least as
long (Appendix~\ref{app:convergence}).
A practical lesson is that a model can show
flat performance over long stretches while still having significant room for
improvement.  We attribute this behavior to the interplay between the evaluator and the policy derived
from it, which evolves as the evaluator improves.

\paragraph{Training targets: sampled vs.\ exact backups.}
Let $x$ be any pseudo-state (Sec.~\ref{sec:background}), and recall that rolling $d \sim
\mathcal{D}$ yields the state $s_{x,d}$, from which any action $a$ leads
deterministically to the next pseudo-state $\tilde{T}(s_{x,d}, a)$.  We use one of two training targets,
both regressing $V(x)$.  The \emph{sampled-Bellman} target estimates the value
from a single dice roll,
\begin{equation}
\label{eq:sampled}
y^{\text{sampled}}(x) \;=\; \textstyle\max_a V\!\bigl(\tilde{T}(s_{x,d}, a)\bigr),
\qquad d \sim \mathcal{D},
\end{equation}
i.e.\ it rolls once, plays greedily, and records the value of the resulting
pseudo-state.  The \emph{exact-Bellman} target instead takes the full expectation
over all 21 rolls,
\begin{equation}
\label{eq:exact-target}
y^{\text{exact}}(x) \;=\; \mathbb{E}_{d \sim \mathcal{D}}\bigl[\, \textstyle\max_a V\!\bigl(\tilde{T}(s_{x,d}, a)\bigr) \,\bigr],
\end{equation}
which is equivalent to the ``1-ply amplified equity''
target of \citet{hilton2019backgammon}.  We use
sampled-Bellman backups for the initial
training (typically $2$--$5.5\times$ faster per episode, depending on hardware)
until a plateau is reached,
then switch to exact-Bellman backups until a second plateau is reached.
Appendix~\ref{app:backup-comparison} compares the two targets from a shared
prob5 checkpoint and discusses their relative cost.

\subsection{The Doubling Cube as an Action-Space Extension}
\label{sec:cube}

Standard engines handle the cube via Janowski's formulas applied on top of a
cubeless evaluator (Sec.~\ref{sec:background}).  We instead treat \emph{cube-offer decisions}, in which the player on roll
must decide whether to double, as ordinary actions in the agent's action
space, which is arguably a more natural RL formulation.  These decisions occur
at the beginning of the side-to-move's turn, provided the cube is centered or
owned by the mover.  As no dice are rolled
during a cube-offer decision, transitions are deterministic and we can apply
Eq.~\eqref{eq:bellman} directly, with the maximization ranging over the cube
actions $\{\texttt{no\_double}, \texttt{offer\_double}\}$.  A cube-decision
state is a pre-roll position, but it is not a pseudo-state in the sense of
Eq.~\eqref{eq:pseudostate}, which describes checker-play decisions.

For cubeful money play we extend the input
to 200 features by adding three
one-hot cube-ownership features (\texttt{cube\_centered}, \texttt{cube\_own},
\texttt{cube\_opp\_own}) and a binary \texttt{is\_cube\_action} flag marking
cube-decision positions.  The flag is $1$ when the on-roll player faces a
cube-offer decision.  It is $0$ when the on-roll player either has no access to
the cube or has already decided not to double, in which case the position calls
for a checker decision.  Note that when \texttt{is\_cube\_action} is $0$, the
dice have not yet been rolled, so the position is a pseudo-state
(Sec.~\ref{sec:background}).  Table~\ref{tab:cube-features} enumerates the
combinations these four features can take and the corresponding decision they
encode.

\begin{table}[h]
\centering
\caption{\textbf{Cube-specific features.}}
\label{tab:cube-features}
{\small
\setlength{\tabcolsep}{5pt}%
\begin{tabular}{@{}cccc l@{}}
\toprule
\texttt{cube\_centered} & \texttt{cube\_own} & \texttt{cube\_opp\_own} &
\texttt{is\_cube\_action} & Decision for the player on roll \\ \midrule
1 & 0 & 0 & 0 & checker decision \\
1 & 0 & 0 & 1 & cube offer (initial cube) \\
\addlinespace
0 & 1 & 0 & 0 & checker decision \\
0 & 1 & 0 & 1 & cube offer (recube) \\
\addlinespace
0 & 0 & 1 & 0 & checker decision \\
0 & 0 & 1 & 1 & not a legal configuration \\
\bottomrule
\end{tabular}}
\end{table}

Because these transitions are deterministic, the sampled and exact
targets of a cube-offer decision coincide:
\begin{equation}
\label{eq:cube-target}
y^{\text{sampled}}(s) \;=\; y^{\text{exact}}(s) \;=\; \textstyle\max_a V\!\bigl(\tilde{T}(s, a)\bigr).
\end{equation}
Thus, during exact-Bellman training, our code takes the more expensive expectation over
dice rolls for checker plays but uses the cheaper update of Eq.~\eqref{eq:cube-target}
for cube-offer decisions.

When a cube is offered, the player faces a take/pass decision.  We resolve this
greedily with respect to the learned value function $V$, taking the cube when the
resulting position is worth more than conceding the current cube value and passing
otherwise.  Thus, the decision is driven by the learned evaluator.  Take and pass could
instead be modeled as explicit learned actions with their own input flag and TD
sample, but this cannot help at the Bellman fixed point.  The take/pass node
involves no dice and, unlike the cube-offer state, is never reached directly
after a checker play.  Its two successors are a terminal state worth exactly
the current cube value and the post-take pseudo-state, which the network
already evaluates.  Its optimal value is therefore the deterministic
$\min(V_{\mathrm{take}}, V_{\mathrm{pass}})$ of
Algorithm~\ref{alg:cube-td}, and a separately learned estimate of it could
only add approximation error and extra computation.

Cubeful backgammon money games are most commonly played with the \emph{Jacoby rule},
under which a gammon or backgammon counts as more than a normal win only after
the cube has been turned (offered and accepted at least once).  During training,
our code handles this directly for the on-roll player during a cube-offer
decision: when the cube is centered, the player offers a double whenever the opponent
will pass.

Algorithm~\ref{alg:cube-td} provides pseudo-code for cube-offer decisions and
their subsequent TD update during self-play.
The single $\max$ in Eq.~\eqref{eq:cube-target} and the explicit $\min$ in
Algorithm~\ref{alg:cube-td} are consistent.  In the equation, the opponent's
take/pass response is folded into the successor value
$V(\tilde{T}(s, \texttt{offer\_double}))$, as the pseudo-state reached after the
\emph{cube offer} is whatever the opponent's equity-minimizing response leads to.
Algorithm~\ref{alg:cube-td} instead resolves that response concretely as
$V_{\mathrm{double}} = \min(V_{\mathrm{take}}, V_{\mathrm{pass}})$.  The factor of
$2$ in the $V_{\mathrm{take}}$ calculation reflects that a take doubles the stake.  It appears as a constant rather than
depending on the current cube level because $V^{\text{model}}(\cdot)$ returns equity
from the mover's view normalized by the current cube value (see ``Cube normalization'' in Sec.~\ref{sec:background}).

\begin{algorithm}[t]
\caption{\textbf{PureTD training for cube decisions}}
\label{alg:cube-td}
\begin{algorithmic}
\STATE {\bfseries Input:} the encoded board $\mathit{board}$ at a cube-offer decision, the cube-ownership binary indicators $\texttt{cube\_centered}$ and $\texttt{cube\_own}$ (exactly one of which is $1$), and the current evaluator $V^{\text{model}}$.
\STATE {\bfseries Notation:} a state concatenates the encoded board with a 4-tuple of binary cube features $(\langle\text{cube centered}\rangle, \langle\text{cube owned by us}\rangle, \langle\text{cube owned by opponent}\rangle, \langle\text{is cube action}\rangle)$, in that order.  $V^{\text{model}}(\cdot)$ is a forward pass of the evaluator (equity from the mover's view normalized by the cube's value), which $\mathrm{TD\_update}$ trains toward the target.
\STATE
\STATE $\mathit{cube} \gets (\texttt{cube\_centered}, \texttt{cube\_own}, 0, 1)$ \quad // \textit{current decision is a cube action}
\STATE $s \gets \mathrm{concat}(\mathit{board}, \mathit{cube})$ \quad // \textit{current state}
\STATE
\COMMENTLINE{Value of not doubling (cube unchanged, next is a checker decision)}
\STATE $\mathit{cube}_{\mathrm{nd}} \gets (\texttt{cube\_centered}, \texttt{cube\_own}, 0, 0)$
\STATE $V_{\mathrm{no\_double}} \gets V^{\text{model}}(\mathrm{concat}(\mathit{board}, \mathit{cube}_{\mathrm{nd}}))$
\STATE
\COMMENTLINE{Value if we double and the opponent takes (opponent now owns the cube, stake doubles)}
\STATE $\mathit{cube}_{\mathrm{dt}} \gets (0, 0, 1, 0)$
\STATE $V_{\mathrm{take}} \gets 2 \cdot V^{\text{model}}(\mathrm{concat}(\mathit{board}, \mathit{cube}_{\mathrm{dt}}))$
\STATE
\COMMENTLINE{If we double and the opponent passes, we win the current cube value (normalized to 1)}
\STATE $V_{\mathrm{pass}} \gets 1$
\STATE
\COMMENTLINE{The opponent takes or passes to minimize our equity}
\STATE $V_{\mathrm{double}} \gets \min(V_{\mathrm{take}}, V_{\mathrm{pass}})$
\STATE $\mathit{opp\_passes} \gets (V_{\mathrm{take}} > V_{\mathrm{pass}})$
\STATE
\COMMENTLINE{We pick the cube action that maximizes our equity under the Jacoby rule}
\IF{$\texttt{cube\_centered} = 1$ and $\mathit{opp\_passes}$}
    \STATE $\mathit{cube\_decision} \gets \texttt{double\_pass}$, \quad $V_{\mathrm{target}} \gets V_{\mathrm{pass}}$ \quad // \textit{Jacoby rule logic}
\ELSIF{($\mathit{opp\_passes}$ and $V_{\mathrm{no\_double}} \leq 1$) or (not $\mathit{opp\_passes}$ and $V_{\mathrm{take}} > V_{\mathrm{no\_double}}$)}
    \IF{$\mathit{opp\_passes}$}
        \STATE $\mathit{cube\_decision} \gets \texttt{double\_pass}$, \quad $V_{\mathrm{target}} \gets V_{\mathrm{pass}}$
    \ELSE
        \STATE $\mathit{cube\_decision} \gets \texttt{double\_take}$, \quad $V_{\mathrm{target}} \gets V_{\mathrm{double}}$
    \ENDIF
\ELSE
    \STATE $\mathit{cube\_decision} \gets \texttt{no\_double}$, \quad $V_{\mathrm{target}} \gets V_{\mathrm{no\_double}}$
\ENDIF
\STATE
\COMMENTLINE{Trains model: TD update of the current state toward the chosen successor value}
\STATE $\mathrm{TD\_update}(\text{input}=s,\ \text{target}=V_{\mathrm{target}})$
\STATE
\IF{$\mathit{cube\_decision} = \texttt{double\_pass}$}
    \STATE the game terminates and we win the current cube value
\ELSE
    \STATE proceed to the checker decision for the same player and board (roll the dice and choose an action)
\ENDIF
\end{algorithmic}
\end{algorithm}

We work through two examples.

\textbf{Example 1.}  Suppose the agent owns the
doubling cube (having previously taken a cube) and is on roll with board
$b$, so it must decide whether to redouble.  The current state $s$ consists of
$b$ together with the cube-ownership one-hot vector
$\mathbf{c} = (\texttt{cube\_centered}, \texttt{cube\_own}, \texttt{cube\_opp\_own})$
and the flag $\texttt{is\_cube\_action}$.  Here $\mathbf{c} = (0, 1, 0)$ (the
agent owns the cube) and $\texttt{is\_cube\_action} = 1$ (the agent faces a cube
decision).  To make a decision, the agent evaluates the possible next
pseudo-states, each of which is either
terminal or shares the same board $b$. Not redoubling
leads to a pseudo-state $s_{\text{no-recube}}$ with $\mathbf{c} = (0, 1, 0)$ and
$\texttt{is\_cube\_action} = 0$.  When redoubling, the on-roll player computes
whether the opponent will take or pass, a decision that is itself greedy with
respect to the model, and
thereby identifies the corresponding next pseudo-state from among the following:
\begin{itemize}
\item Redoubling followed by a take leads to a pseudo-state $s_{\text{recube}}$ with
$\mathbf{c} = (0, 0, 1)$ and $\texttt{is\_cube\_action} = 0$ (after the redouble
the opponent owns the cube, and the agent is back to a checker decision).\footnote{The
opponent passes precisely when $2V^{\text{model}}(s_{\text{recube}}) > 1$.}
\item Redoubling followed by a pass leads to a terminal state in which the agent
wins the current cube value.
\end{itemize}
Now suppose the agent redoubles and the opponent takes.  The state
transitions from $s$ to $s_{\text{recube}}$, and we store a TD update training
$V(s)$ toward $V(s_{\text{recube}})$, that is, $V^{\text{model}}(s)$ toward
$2V^{\text{model}}(s_{\text{recube}})$ (the stake doubles after a take).  If instead the opponent passes, the game
transitions to a terminal state in which the on-roll agent wins, again
triggering a TD update for $s$.  Figure~\ref{fig:cube-example} diagrams this
worked example.

\begin{figure}[t]
\centering
\begin{tikzpicture}[
  >={Stealth[length=2mm]},
  st/.style={draw, rounded corners, align=center, font=\footnotesize, inner sep=3pt, text width=31mm, fill=blue!5},
  tm/.style={draw, rounded corners, align=center, font=\footnotesize, inner sep=3pt, text width=31mm, fill=black!8},
  op/.style={draw, diamond, aspect=1.5, align=center, font=\scriptsize, inner sep=0pt, fill=orange!15},
  al/.style={font=\scriptsize, inner sep=1.5pt, fill=white},
  fwd/.style={->, thick},
  td/.style={->, dashed, gray!55!black},
]
\node[st]  (s)    at (0,0)      {\textbf{$s$: cube decision}\\agent owns cube\\$\mathbf{c}{=}(0,1,0)$\\\texttt{is\_cube\_action}${=}1$};
\node[st]  (nod)  at (7.5,2.3)  {$s_{\text{no-recube}}$\\(agent's checker play)\\$\mathbf{c}{=}(0,1,0)$\\\texttt{is\_cube\_action}${=}0$};
\node[op]  (op)   at (4.3,-1.6) {opp.\\take/pass};
\node[st]  (rec)  at (7.5,0.1)  {$s_{\text{recube}}$\\(opp.\ owns cube)\\$\mathbf{c}{=}(0,0,1)$\\\texttt{is\_cube\_action}${=}0$};
\node[tm]  (term) at (7.5,-2.7) {\textbf{terminal}\\agent wins cube value};

\draw[fwd] (s)  -- node[al,above,sloped]{\texttt{no\_double}} (nod);
\draw[fwd] (s)  -- node[al,below,sloped,font=\tiny]{\texttt{offer\_double}} (op);
\draw[fwd] (op) -- node[al,above,sloped,font=\tiny]{take} (rec);
\draw[fwd] (op) -- node[al,below,sloped,font=\tiny]{pass} (term);

\draw[td] (nod)  to[bend right=18] node[al,above]{TD} (s);
\draw[td] (rec)  to[bend right=7]  node[al,above]{TD} (s);
\draw[td] (term) to[bend right=26] node[al,below,pos=0.8]{TD} (s);
\end{tikzpicture}
\caption{\textbf{Example 1 as a sequence of state transitions.}  Dashed arrows
mark the TD update that backs the realized successor value up to $V(s)$.}
\label{fig:cube-example}
\end{figure}

\textbf{Example 2.}  Suppose the cube is centered and the agent on
roll faces a cube decision: $\mathbf{c} = (1, 0, 0)$ and
$\texttt{is\_cube\_action} = 1$.  Suppose the agent chooses \texttt{no\_double}.
The state transitions to $s_{\text{no-double}}$ with the same board,
$\mathbf{c} = (1, 0, 0)$, and $\texttt{is\_cube\_action} = 0$: the cube remains centered, and the agent
now makes its checker play.  Even though no offer was made, this is still a
real state transition, and we generate a TD update training $V(s)$ toward
$V(s_{\text{no-double}})$.  The cube-decision state and the subsequent
checker-play pseudo-state are distinct network inputs (where only the \texttt{is\_cube\_action}
flag differs).

Our treatment of cube decisions follows the same spirit as
\citet{lin2020cube} but corrects a key conceptual flaw.  Lin's network
encodes cube ownership but has no input indicating whether it is evaluating
a cube decision or a checker play.  Thus Lin's network cannot distinguish
the state during a cube-offer decision from the post-no-double
pseudo-state.  Lin generates a training sample for the cube decision when
an offered cube is taken or passed.  When the
agent chooses not to double (as in Example~2), the cube-decision state is
trained only after the subsequent checker play, and a sample for the
no-double decision itself would be a no-op.  We
instead generate a TD training sample for every cube-offer decision (\texttt{offer\_double} or \texttt{no\_double}) and
add \texttt{is\_cube\_action}, allowing the network to distinguish a
cube-decision query from a checker-play query at the same board position.

In an early implementation, similar to Lin, we omitted \texttt{is\_cube\_action},
forcing the cube-decision state and the post-no-double pseudo-state to share a value.  Play was
strong, but the agent failed to double in clear double/pass positions: the
cube-decision value is exactly $1$ (double, opponent passes), while the
post-no-double value is the equity of continued play, typically much less than
$1$.  The decision rule needs both, but a single shared estimate cannot represent
them.  For a full ablation study of our approach compared to Lin's, see
Sec.~\ref{sec:cube-ablation}.

%%%%%%%%%%%%%%%%%%%%%%%%%%%%%%%%%%%%%%%%%%%%%%%%%%%%%%%%%%%%%%%%%%%%%%%%%%%%%%%%
\section{Experiments}
\label{sec:experiments}

\subsection{Baselines and Metrics}
\label{sec:baselines}

\paragraph{Baselines.}
We compare against two open-source engines, to our knowledge the strongest
available as of July 15, 2026 (see Sec.~\ref{sec:background}).  Our primary
baseline is \gnubg{}~\citep{gnubg}.
To play against \gnubg{}, we run its binary as a separate process and
drive an agent through its external-player socket interface, so every checker
move and cube decision is produced by \gnubg{} itself.  Also included is Open
Sage~\citep{bgsage}.  For
1-ply, both baselines use their developers' default move pruning throughout.  Disabling it
has a very minor impact on results.  Analysis settings and further details
are in Appendix~\ref{app:repro}.

PureTD does not use bearoff databases in our experiments, except against our
Janowski-based baseline in
Sec.~\ref{sec:cube-ablation}, where both sides use them.  Unless stated
otherwise, games use independent dice with no variance
reduction.

\paragraph{Metrics.}
The ideal metric for comparing backgammon engines is head-to-head play over
many games, and we report the result in mEq/game.  Because individual game outcomes are noisy, precise comparisons
can require large samples.  Appendix~\ref{app:convergence} describes a standard
method for variance reduction based on a luck
adjustment~\citep{dahl2001variance,montgomery2000variance}.

In the cubeful setting, we
also report an offline analysis of each engine's play by XG~\citep{xg},
summarized by its \emph{Performance Rating} (PR), an error rate for which
lower is better.  For human reference, the Backgammon Masters Awarding Body~\citep{bmab}
awards its top titles by XG's PR: a PR of $4$ or lower earns ``Grand
Master'' and a PR of $2.5$ or lower ``Super Grand Master'' (held by only two
living players).  Play below a PR of roughly $2$ is generally regarded as
super-human.

\paragraph{XG++ Performance Rating.}
XG++ is eXtreme Gammon's strongest standard analysis level, and we
use it to compute the Performance Rating,
\[
\mathrm{PR} \;=\; 500 \times \frac{\text{total equity lost}}{\text{number of unforced decisions}},
\]
where equity lost is computed against the XG++ evaluation, a
proprietary truncated-rollout method.  Thus a PR of $1.0$ corresponds to an
average loss of $0.002$ equity per decision, and a PR of $0$ means XG++ found no errors.
Where we report \emph{checker PR} and \emph{cube PR}, they are the same
quantity restricted to checker-play decisions and to cube decisions,
respectively.  Unless stated otherwise, each PR is computed from $1{,}000$
self-play games.

\paragraph{Confidence intervals.}
All reported intervals are $95\%$.  Head-to-head equity and XG++ PR use
game-level percentile bootstraps ($B=10{,}000$ for head-to-head equity and
$B=100{,}000$ for XG++ PR), and both players of a self-play game enter a resample
together, since the two sides of the same game are not independent.  The
throughput intervals of Sec.~\ref{sec:cost} are not bootstraps.  Each is a
$t$-interval over eight independent benchmark runs (sweeps, Sec.~\ref{sec:cost}), because
run-to-run variation exceeds the within-run sampling error.

\subsection{Experimental Results}
\label{sec:results}

\paragraph{Cubeless money.}
Modern backgammon engines such as \gnubg{}, Open Sage, BGBlitz~\citep{bgblitz}, and
XG~\citep{xg, xgmanual} train a
cubeless money evaluator (for money games without a doubling cube) that
predicts nested outcome probabilities (win, gammon or better, and
backgammon).  For a more level
comparison of the evaluator alone, we also train such a model, which we call
\emph{prob5}.  Cubeless results labeled PureTD use this
prob5 model.

The prob5 model uses the same 196-feature board encoding as our cubeful model
and, following standard practice, has five probability outputs:
$P(\mathrm{win})$, $P(\mathrm{win\ gammon})$, $P(\mathrm{win\ backgammon})$,
$P(\mathrm{lose\ gammon})$, and $P(\mathrm{lose\ backgammon})$, from the
side-to-move's perspective.  The five outputs are independent sigmoids, trained
by a slight modification of the TD procedure of Sec.~\ref{sec:architecture}.
Action selection, in both training and evaluation, uses the scalar equity
\[
\mathrm{Eq} = 2\,P(\mathrm{win}) - 1 + P(\mathrm{win\ gammon}) + P(\mathrm{win\ backgammon})
    - P(\mathrm{lose\ gammon}) - P(\mathrm{lose\ backgammon}).
\]
Each output is then regressed
toward its own bootstrapped value at the position reached by this
equity-greedy action, with a terminal reward of $1$ for the corresponding
outcome and $0$ otherwise.  For example, writing $\hat{P}_{\mathrm{win}}$ for the
win-probability output (converted, like $V$, to the maximizer's perspective), its sampled- and
exact-Bellman targets are
\[
y^{\text{sampled}}_{\mathrm{win}}(x) = \hat{P}_{\mathrm{win}}\bigl(\tilde{T}(s_{x,d}, a^\star_d)\bigr),
\qquad
y^{\text{exact}}_{\mathrm{win}}(x) = \mathbb{E}_{d \sim \mathcal{D}}\Bigl[\hat{P}_{\mathrm{win}}\bigl(\tilde{T}(s_{x,d}, a^\star_d)\bigr)\Bigr],
\]
where $a^\star_d = \arg\max_a \mathrm{Eq}\bigl(\tilde{T}(s_{x,d}, a)\bigr)$ and, for the
sampled target, $d \sim \mathcal{D}$ is a single roll.

At matched
0-ply our prob5 model is substantially stronger than both baselines
(Table~\ref{tab:cubeless-results}).  PureTD's advantage at 0-ply persists even
when the baselines are given an extra ply of search.

\begin{table}[h]
\centering
\caption{\textbf{Cubeless-money strength.}  Our prob5 model (four hidden layers, $[512{,}512{,}256{,}256]$,
562k parameters) versus the baselines.  Each row gives that
engine's head-to-head equity against our prob5 model at 0-ply, over $10^7$ games
(negative indicates a loss to our model).  Brackets give
95\% confidence intervals.}
\label{tab:cubeless-results}
\begin{tabular}{@{}lr@{}}
\toprule
Engine & vs.\ PureTD 0-ply (mEq/game) \\ \midrule
\gnubg{}, 0-ply      & $-45.8$ $[-46.7, -45.0]$ \\
\gnubg{}, 1-ply      & $-12.1$ $[-12.9, -11.2]$ \\
Open Sage, 0-ply     & $-36.8$ $[-37.7, -35.9]$ \\
Open Sage, 1-ply     & $-5.9$ $[-6.8, -5.1]$ \\
\bottomrule
\end{tabular}
\end{table}

\paragraph{Cubeful money.}
We now turn to our main model, which learns to play end-to-end with the doubling
cube.  In addition to head-to-head results, we perform offline evaluation
of our model and the baselines with XG analysis.  The offline analysis
was performed on self-play games (same engine playing both sides), but we
found it to be generally robust when the analysis was instead performed on games
between separate agents (Appendix~\ref{app:mixed-play}).  All
cubeful money games in this paper, in training and evaluation alike, use the
Jacoby rule (Sec.~\ref{sec:cube}).

During play,
a game occasionally reaches a complicated back game that neither engine
understands, and both sides cube aggressively, leading to runaway cube
values.  Such games are extremely rare, roughly two per million, but
have an outsized effect on the mean and variance.  Thus, to obtain tight
confidence intervals, per-game equity was capped at $\pm 128$ in all
head-to-head results in this paper.  Note that in money play,
once the cube has been turned, optimal play is the same regardless of the
cube's value.  Capping the equity of these runaway games therefore does not
conceal a major deficiency in play, except possibly in the complicated back
games themselves.  Uncapped means for our primary head-to-head results, and
means under higher equity caps, are in Appendix~\ref{app:caps}.

Similar to the cubeless-money results, our cubeful model is substantially
stronger than both baselines (Table~\ref{tab:money-results}) and remains stronger
even when opponents are allowed a narrow search: PureTD
at 0-ply wins $27.2$ mEq/game versus \gnubg{} at 1-ply and $23.8$ versus Open
Sage at 1-ply.

\begin{table}[h]
\centering
\caption{\textbf{Cubeful-money strength.}  Our production model (four hidden layers, $[512{,}512{,}256{,}256]$, 563k
parameters) versus the baselines.  The XG++ PR column uses $1{,}000$
self-play games per row.  The last column is each engine's
head-to-head equity against our model at 0-ply (negative indicates a loss to
our model) over $10^7$ games ($2 \times 10^6$ for the Open Sage rows).
Brackets give $95\%$ confidence intervals.}
\label{tab:money-results}
\begin{tabular}{@{}lrr@{}}
\toprule
Engine & XG++ PR & vs.\ PureTD 0-ply (mEq/game) \\ \midrule
PureTD, 0-ply & $0.80$ $[0.76, 0.84]$ & ---                     \\
\gnubg{}, 0-ply & $2.21$ $[2.12, 2.30]$ & $-85.2$ $[-87.0, -83.5]$ \\
\gnubg{}, 1-ply & $1.25$ $[1.20, 1.32]$ & $-27.2$ $[-29.0, -25.3]$ \\
Open Sage, 0-ply     & $2.11$ $[2.02, 2.20]$ & $-79.5$ $[-83.8, -75.2]$ \\
Open Sage, 1-ply     & $1.24$ $[1.18, 1.29]$ & $-23.8$ $[-28.2, -19.5]$ \\
\bottomrule
\end{tabular}
\end{table}

\FloatBarrier
\subsection{Computational Cost}
\label{sec:cost}

We report raw evaluation throughput, measured on the CPU (GPUs were used during
training but not for any results in this section).  Our implementation was not
optimized for speed beyond calling PyTorch's C++ library from our C game engine.  In
Appendix~\ref{app:optimized}, however, we provide preliminary
results from basic optimizations that give, at 0-ply, about a $3$--$8\times$ speedup
on checker decisions and $8$--$11\times$ on cube decisions.  We
use the fixed-position benchmark described below, reporting throughput on two
machines, an AMD Ryzen~5 9600X (Zen~5) and an AMD Ryzen~9 5950X (Zen~3).
Appendix~\ref{app:repro} details each engine's
evaluation path and the benchmark scripts that reproduce these numbers.
Each benchmarked configuration is run eight times (each run is termed a sweep), and the results
are averaged.

\paragraph{Fixed-position benchmark.}
Each engine is run on a single thread over the same fixed pools of decision
points sampled from completed games ($8{,}000$ cubeless positions, and
$10{,}000$ checker and $6{,}000$ cube decisions for cubeful play, with details in
Appendix~\ref{app:repro}).

Tables~\ref{tab:cubeless-cost} (cubeless) and~\ref{tab:money-cost} (cubeful
money) report cost and strength together, with throughput rounded to three
significant figures (Appendix~\ref{app:bench-tables} gives the full values with
confidence intervals).  At 0-ply our code is about $3$--$8\times$ slower than
\gnubg{} and Open Sage on checker decisions, partly because our networks are
larger and partly because both baselines have optimized evaluation code.
On AMD CPUs, MKL, which PyTorch uses for matrix multiplication, does not
select its AVX-512 kernels, so the PyTorch checker-decision rows gain little
from the move to Zen~5.

At 1-ply, the comparison is further complicated by move pruning.  Each engine
runs with its developers' default settings.  In cubeless play, Open Sage keeps
the top five moves within $0.08$ equity by its 0-ply evaluation, and when an
opponent roll has more than $20$ legal replies it keeps the $15$ best under
the linear evaluator PubEval~\citep{tesauro1993pubeval}.  \gnubg's
filter is looser (up to eight moves within $0.16$), and it selects
opponent replies with dedicated pruning networks.  Open Sage's
cubeful checker play, as provided by its analyzer and C API, has no pruning,
while its cube decisions rank the moves of the player on roll with PubEval and
keep the $15$ best when more than $16$ are legal.  Open Sage also has a pruned
cubeful 1-ply search, which its rollouts use for checker moves.  We time it with
Open Sage's default move filter (the top five moves within $0.08$ equity by
0-ply cubeful evaluation), and it chooses the same move as the
full-width search on $98\%$ of the benchmark positions.  Table~\ref{tab:money-cost} times it as a
separate row, without a strength entry because no games were played with it.
We view this pruned row as more representative of Open Sage's speed.
For \gnubg{} (cubeless and cubeful) and Open Sage (cubeless), we found that
pruning had a very minor impact on playing strength
(Appendix~\ref{app:repro}).

Taken together, these results point to a favorable cost versus strength trade-off for
PureTD: our 0-ply network is stronger than both \gnubg{}
and Open Sage even when they are granted a 1-ply search, yet evaluates
at least $3.5\times$ faster than that search.
Table~\ref{tab:money-cost} also includes a row for \gnubg{} at 2-ply under
its default 2-ply pruning, the shallowest search depth at which \gnubg{} beats
PureTD at 0-ply in cubeful play.

\begin{table}[h]
\centering
\caption{\textbf{Cubeless-money cost versus strength.}  Strength is each baseline's
head-to-head equity (mEq/game) against PureTD 0-ply.  Cost
is single-thread checker-decision throughput (decisions/s) under the
fixed-position benchmark.  After each dec/s we provide the
per-decision cost factor relative to PureTD ($2\times$ means half as fast as
PureTD).}
\label{tab:cubeless-cost}
{\small
\setlength{\tabcolsep}{4pt}%
\begin{tabular}{@{}llrrr@{}}
\toprule
Engine & Depth & Equity vs.\ PureTD (mEq/game) & dec/s (Zen 3) & dec/s (Zen 5) \\ \midrule
PureTD & 0-ply & --- & $4{,}550$ & $5{,}090$ \\
\gnubg{} & 0-ply & $-45.8$ & $34{,}600$ ($0.13\times$) & $42{,}700$ ($0.12\times$) \\
Open Sage & 0-ply & $-36.8$ & $16{,}900$ ($0.27\times$) & $19{,}900$ ($0.26\times$) \\
\addlinespace
\gnubg{} & 1-ply & $-12.1$ & $1{,}000$ ($4.5\times$) & $1{,}340$ ($3.8\times$) \\
Open Sage & 1-ply & $-5.9$ & $824$ ($5.5\times$) & $1{,}090$ ($4.7\times$) \\
\bottomrule
\end{tabular}}
\end{table}

\begin{table}[h]
\centering
\caption{\textbf{Cubeful-money cost versus strength.}  Strength is each
baseline's head-to-head equity (mEq/game) against PureTD 0-ply.
Cost is single-thread throughput (decisions/s) under the fixed-position
benchmark, for checker and cube decisions separately.  The \gnubg{} 2-ply result uses $2 \times 10^6$ games ($95\%$
interval $[+16.6, +24.8]$).
Each dec/s entry is followed by the per-decision cost factor relative to
PureTD, as in Table~\ref{tab:cubeless-cost}.
$^{\dagger}$Open Sage's 1-ply row is the full-width cubeful search of its
analyzer and C API, with which the strength games were played.  The pruned
row times the pruned cubeful 1-ply search its rollouts use, run with the default
filter, as described above.}
\label{tab:money-cost}
{\small
\setlength{\tabcolsep}{4pt}%
\resizebox{\textwidth}{!}{%
\begin{tabular}{@{}llrrrrr@{}}
\toprule
 & & & \multicolumn{2}{c}{Checker dec/s} & \multicolumn{2}{c}{Cube dec/s} \\
\cmidrule(lr){4-5}\cmidrule(lr){6-7}
Engine & Depth & Equity vs.\ PureTD (mEq/game) & Zen 3 & Zen 5 & Zen 3 & Zen 5 \\ \midrule
PureTD & 0-ply & --- & $5{,}050$ & $5{,}640$ & $13{,}000$ & $24{,}500$ \\
\gnubg{} & 0-ply & $-85.2$ & $34{,}200$ ($0.15\times$) & $41{,}600$ ($0.14\times$) & $605{,}000$ ($0.02\times$) & $715{,}000$ ($0.03\times$) \\
Open Sage & 0-ply & $-79.5$ & $16{,}700$ ($0.30\times$) & $20{,}300$ ($0.28\times$) & $276{,}000$ ($0.05\times$) & $349{,}000$ ($0.07\times$) \\
\addlinespace
\gnubg{} & 1-ply & $-27.2$ & $1{,}090$ ($4.6\times$) & $1{,}440$ ($3.9\times$) & $3{,}170$ ($4.1\times$) & $4{,}110$ ($6.0\times$) \\
Open Sage & 1-ply & $-23.8$ & $40.2$ ($126\times$) & $52.0$ ($108\times$) & $1{,}460$ ($8.8\times$) & $1{,}870$ ($13.1\times$) \\
Open Sage, pruned$^{\dagger}$ & 1-ply & --- & $910$ ($5.6\times$) & $1{,}200$ ($4.7\times$) & --- & --- \\
\gnubg{} & 2-ply & $+20.7$ & $56.5$ ($89\times$) & $75.3$ ($75\times$) & $188$ ($69\times$) & $248$ ($99\times$) \\
\bottomrule
\end{tabular}}}
\end{table}

\FloatBarrier
\subsection{Effect of Search}
\label{sec:search}

Although the main goal of this paper is to examine RL techniques in the
0-ply setting, it is natural to wonder how engines based
on these models improve with search.  Under matched 1-ply and 2-ply look-ahead,
our cubeful model's offline error rate continues to fall as
the search depth increases (Table~\ref{tab:search-results}), and at every
matched depth it is lower than both baselines'.  As in
our previous strength comparisons, the baselines use their default 1-ply move
pruning.  For the
XG++ PR analysis, every engine at 2-ply evaluates every legal root move, and only the
opponent's replies are pruned: \gnubg{} with its default pruning networks, Open Sage
with its own default reply selection, and PureTD by applying 1-ply analysis
only to the opponent's top reply as ranked at 0-ply.  This differs from the 2-ply
configuration used for games and timing in Sec.~\ref{sec:cost}, which uses
\gnubg's default root move filter.  Playing games is slow at these depths, so this
section relies mainly on offline XG analysis, though
Table~\ref{tab:money-cost} reports a 2-ply head-to-head result against
\gnubg{} alongside its per-decision cost.

\begin{table}[h]
\centering
\caption{\textbf{Effect of look-ahead depth on cubeful money playing strength.}
XG++ PR at matched search depth, with $1{,}000$ self-play games per cell and
95\% confidence intervals in brackets.}
\label{tab:search-results}
{\small
\begin{tabular}{@{}lrrr@{}}
\toprule
Engine & 0-ply & 1-ply & 2-ply \\ \midrule
PureTD            & $0.80$ $[0.76, 0.84]$ & $0.40$ $[0.37, 0.44]$ & $0.29$ $[0.26, 0.33]$ \\
\gnubg{}          & $2.21$ $[2.12, 2.30]$ & $1.25$ $[1.20, 1.32]$ & $0.51$ $[0.48, 0.55]$ \\
Open Sage         & $2.11$ $[2.02, 2.20]$ & $1.24$ $[1.18, 1.29]$ & $0.40$ $[0.37, 0.43]$ \\
\bottomrule
\end{tabular}}
\end{table}

\subsection{Comparison of Cube-Handling Approaches}
\label{sec:cube-ablation}

This section compares our cube handling with two alternatives: the standard
practice of pairing a cubeless evaluator with Janowski's formulas, and the
formulation of \citet{lin2020cube} (the most closely related prior work).

Our implementation of the Janowski approach is outlined in
Appendix~\ref{app:janowski} and follows the logic of \gnubg's
implementation closely, except for bearoff positions, which are described
below.  For the cube-life index, we compared several fixed values and found
that $x = 0.76$ was the best of $\{0.60, 0.68, 0.70, 0.76\}$ and its performance was
within confidence bounds of \gnubg's default adaptive method, which at 0- and 1-ply uses $0.76$ for contact positions
and a pip-count-based heuristic value for race (non-contact) positions.
Since choosing $x$ properly in bearoff
positions is a known difficulty, both agents in this comparison use the
same bearoff databases (Appendix~\ref{app:janowski}).

In Table~\ref{tab:janowski}, PureTD's strength is measured by head-to-head play
against Prob5Jan, our prob5 model combined with Janowski-style cube handling.
Prob5Jan uses \gnubg's adaptive method for $x$ in all reported results.
Since the two models are similar in size and training
method, this provides a direct comparison of our approach with standard
practice.  As the table shows, learning cubeful play directly is superior,
winning about $15$ mEq/game.  PureTD also has the lower XG++ PR, with most of
the difference in cube decisions.
To understand where the advantage comes from, we generated $20{,}000$ games of
PureTD self-play, asked both engines for their choice at every decision point,
and used \gnubg{} at 4-ply to estimate the equity won or lost per game on the
disagreements.  Cube actions account for more of the advantage
than checker play ($5.4$ versus $4.0$ mEq/game,
Appendix~\ref{app:disagreements}), but the considerable checker-play share is
consistent with the cube's influence on checker play.
Finally, Figure~\ref{fig:disagreements} shows one checker play and one cube
decision where PureTD and Prob5Jan disagree, with XG rollout arbitration
favoring PureTD in both.

The two networks in Table~\ref{tab:janowski} are our production models.  Each
was trained in many successive stages, without a fixed recipe, until its
performance appeared to have converged, and the cubeful model received about
$50\%$ more training episodes than the prob5 model.  The recipes in
Appendix~\ref{app:retrain} train both models from scratch to within about
$1$ mEq/game of the production models' strength.
Appendix~\ref{app:convergence} studies continued training.  There we repeat
this comparison with models trained from scratch by the recipes and then
trained further, giving the prob5 model at least as many training episodes as
the cubeful model (Table~\ref{tab:janowski-budgets}).  Both models improved
with the further training, but the advantage of the cubeful model was nearly
unchanged.

\begin{table}[h]
\centering
\caption{\textbf{Learned versus heuristic cube handling.}  PureTD versus
Prob5Jan, both at 0-ply.  The two networks are trained separately but have
the same hidden-layer sizes, $[512{,}512{,}256{,}256]$.  Their architectures differ
only in the four cube input features and the output head.  The head-to-head column uses $10^7$ games.
PureTD's PR games used no bearoff databases, as in
Table~\ref{tab:money-results}, while Prob5Jan's PR games used them, as in its
head-to-head play.  Brackets give $95\%$ confidence intervals.}
\label{tab:janowski}
{\small
\setlength{\tabcolsep}{4pt}%
\begin{tabular}{@{}lrrrr@{}}
\toprule
Model & XG++ PR & Checker PR & Cube PR & vs.\ PureTD, 0-ply (mEq/game) \\ \midrule
PureTD   & $0.80$ $[0.76, 0.84]$ & $0.85$ $[0.80, 0.89]$ & $0.56$ $[0.46, 0.67]$ & --- \\
Prob5Jan & $1.13$ $[1.06, 1.22]$ & $1.05$ $[0.97, 1.14]$ & $1.58$ $[1.36, 1.83]$ & $-14.6$ $[-16.3, -12.8]$ \\
\bottomrule
\end{tabular}}
\end{table}

\begin{figure}[!htb]
\centering
\begin{subfigure}[t]{0.46\textwidth}
\centering
\includegraphics[width=\textwidth]{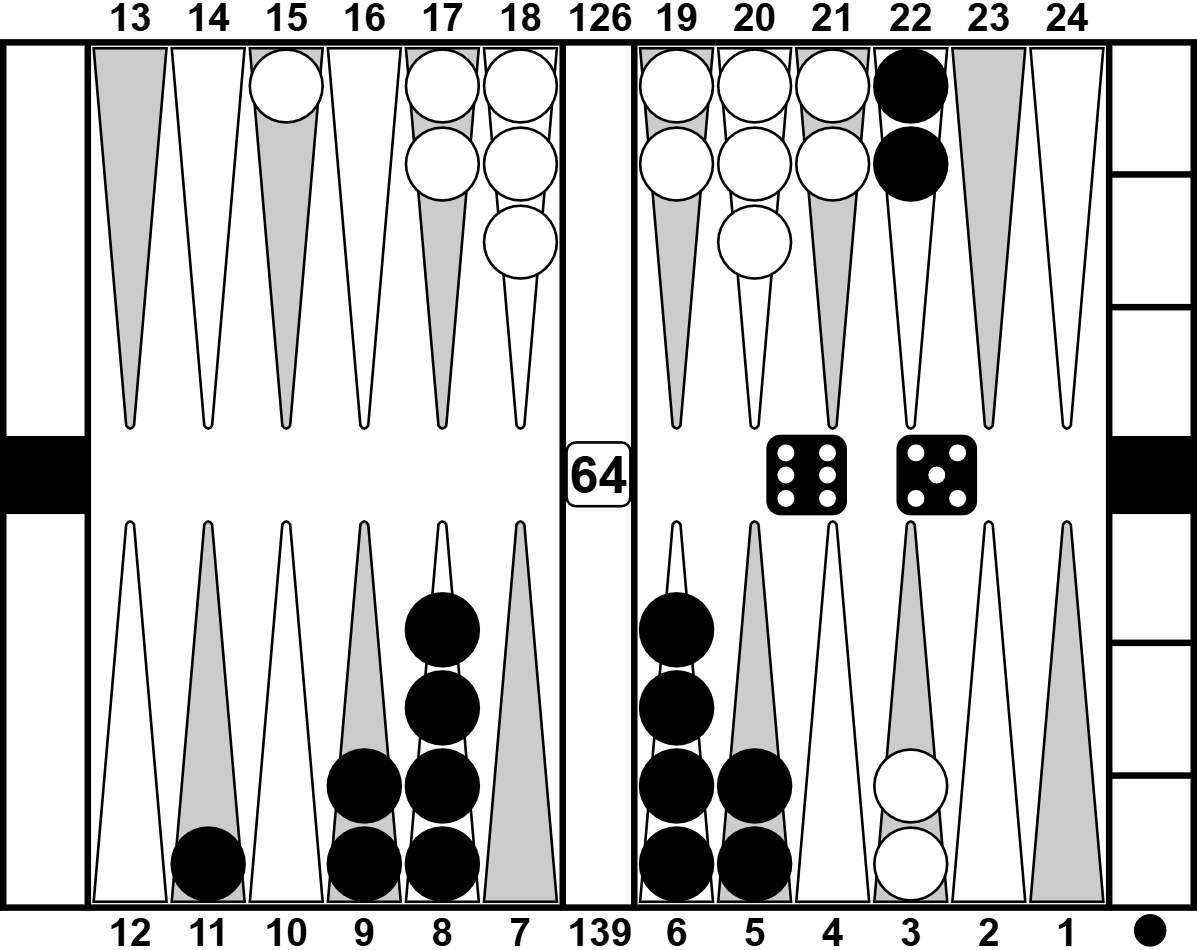}
\caption{Checker play, Black to play 6-5.\\
{\scriptsize\nolinkurl{XGID=---bBCBCB-A---a-bd-db-B---:0:0:-1:65:0:0:1:0:10}}}
\label{fig:disagree-checker}
\end{subfigure}
\hfill
\begin{subfigure}[t]{0.46\textwidth}
\centering
\includegraphics[width=\textwidth]{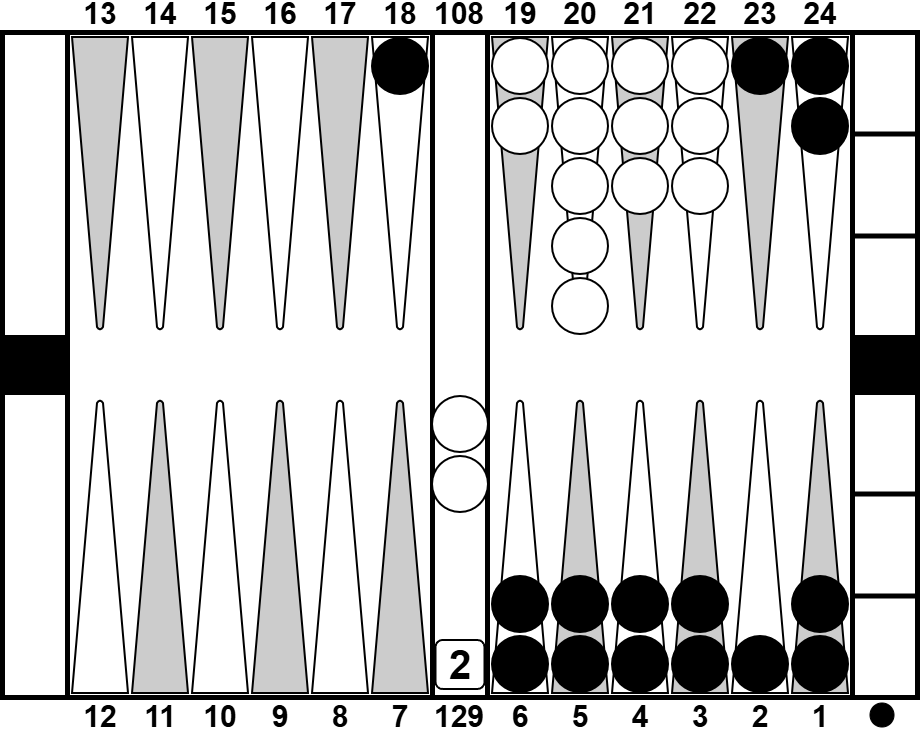}
\caption{2-cube held, redouble decision, Black on roll.\\
{\scriptsize\nolinkurl{XGID=-baCCEBa-----------bbbbabB:1:-1:-1:00:0:0:1:0:10}}}
\label{fig:cube-pos2}
\end{subfigure}

\vspace{8pt}
{\small
\setlength{\tabcolsep}{6pt}%
\begin{tabular}{@{}l lcc lcc@{}}
\toprule
& \multicolumn{3}{c}{(a) Checker play} & \multicolumn{3}{c}{(b) Redouble decision} \\
\cmidrule(lr){2-4}\cmidrule(lr){5-7}
 & Chosen & \texttt{11/5 6/1} & \texttt{22/11} & Chosen & No redbl.\ & Redbl., take \\
\midrule
XG rollout & \texttt{11/5 6/1} & $-0.227$ & $-0.304$ & Redouble, take & $+0.786$ & $+0.828$ \\
PureTD     & \texttt{11/5 6/1} & $-0.249$ & $-0.270$ & Redouble, take & $+0.747$ & $+0.794$ \\
Prob5Jan   & \texttt{22/11}    & $-0.249$ & $-0.219$ & No redouble    & $+0.708$ & $+0.656$ \\
\bottomrule
\end{tabular}}
\caption{\textbf{Two positions where learned and heuristic cube handling
differ.}  In (a) the choice is between playing safe with
\texttt{11/5 6/1}, the best cubeful play, and running off the anchor with
\texttt{22/11}, the best DMP play (winning about $0.4$ percentage points more games but losing
$19$ percentage points more gammons according to the rollout).  In (b) Prob5Jan declines to redouble despite the opponent's two checkers on the
bar, costing $0.042$ equity.}
\label{fig:disagreements}
\end{figure}

To measure the impact of our two changes to the formulation of
\citet{lin2020cube} (Sec.~\ref{sec:cube}), we trained a variant of our cubeful
money model with both removed, using the same recipe as our retrained
cubeful model (Table~\ref{tab:retrain-cubeful} in Appendix~\ref{app:retrain}).
As shown in Table~\ref{tab:cube-ablation}, in head-to-head play at matched
0-ply over $2 \times 10^7$ games, the production model wins $7.6$ mEq/game
against the ablation and cuts
the XG++ PR by $25\%$, with the main improvement in cube decisions.

\begin{table}[h]
\centering
\caption{\textbf{Cube-handling ablation.}  PureTD versus the Lin-style
ablation, both at 0-ply.  The head-to-head column uses $2 \times 10^7$ games.
Brackets give $95\%$ confidence intervals.}
\label{tab:cube-ablation}
{\small
\setlength{\tabcolsep}{4pt}%
\begin{tabular}{@{}lrrrr@{}}
\toprule
Model & XG++ PR & Checker PR & Cube PR & vs.\ PureTD, 0-ply (mEq/game) \\ \midrule
PureTD             & $0.80$ $[0.76, 0.84]$ & $0.85$ $[0.80, 0.89]$ & $0.56$ $[0.46, 0.67]$ & --- \\
Lin-style ablation & $1.07$ $[1.01, 1.13]$ & $1.01$ $[0.95, 1.06]$ & $1.39$ $[1.19, 1.60]$ & $-7.6$ $[-8.8, -6.4]$ \\
\bottomrule
\end{tabular}}
\end{table}

\FloatBarrier
\subsection{Impact of Network Size}
\label{sec:network-size}

We experimented with several network sizes and report their strength and
cost relative to the four-hidden-layer models used throughout the paper
(Table~\ref{tab:capacity} and Figure~\ref{fig:capacity-cost}).  A complete engine would want to optimize the network
size jointly with the run-time (expectimax) search settings to achieve maximum
strength within a fixed time budget.
Each model with five or more hidden layers was trained until a final leg of
at least $10$M episodes changed
head-to-head strength by less than $0.5$ mEq/game.  The smaller models, with
one to three hidden layers, were
trained on fixed schedules and not to convergence.
Unlike the retrained models of Appendix~\ref{app:retrain}, none of these
were trained with a tested recipe, and a small to moderate further
improvement is likely, as continued training showed for the
four-layer models (Appendix~\ref{app:convergence}).

\begin{table}[!htb]
\centering
\caption{\textbf{Strength across network sizes.}  Each row is one trained
model, with its hidden-layer widths and parameter count.  Column ``vs.\ production''
is the model's head-to-head equity (mEq/game) at 0-ply against the
production model of the same family.  Labels ``1 wide'' and ``2 wide'' indicate one or
two leading hidden layers widened to $768$ units.  Cubeful entries are capped ($\pm 128$) means
with Jacoby on.  The last column gives single-thread checker
decisions/s at 0-ply on Zen~5 under the fixed-position benchmark of
Sec.~\ref{sec:cost}, followed by the per-decision cost factor relative to
the production model, as in Table~\ref{tab:cubeless-cost}.  Decisions/s were
timed in a separate set of sweeps from Tables~\ref{tab:cubeless-cost}
and~\ref{tab:money-cost} (eight per host, same protocol).  The production
models' rates differ from those tables by under $1\%$, and each cost factor
is relative to the production model in the same sweeps.  Brackets give
95\% confidence intervals.  Each head-to-head comparison uses $10^7$ games.}
\label{tab:capacity}
{\footnotesize
\setlength{\tabcolsep}{4pt}%
\resizebox{\textwidth}{!}{%
\begin{tabular}{@{}llrrr@{}}
\toprule
Model & Hidden layers & Params & vs.\ production (mEq/game) & Checker dec/s (Zen 5) \\ \midrule
\multicolumn{5}{@{}l}{\emph{Cubeful money}} \\
1-layer  & $[512]$                     & 103k & $-158$ $[-160, -155]$ & $28{,}300$ ($0.20\times$) \\
2-layer  & $[512,512]$                 & 366k & $-25$ $[-27, -23]$    & $8{,}580$ ($0.66\times$) \\
3-layer  & $[512,512,256]$             & 497k & $-5$ $[-6, -3]$       & $6{,}390$ ($0.88\times$) \\
4-layer (production) & $[512,512,256,256]$ & 563k & ---                     & $5{,}650$ \\
5-layer  & $[512,512,256,256,256]$     & 629k & $+5$ $[+3, +7]$       & $5{,}060$ ($1.12\times$) \\
5-layer, 1 wide & $[768,512,256,256,256]$ & 811k & $+8$ $[+7, +10]$      & $3{,}970$ ($1.42\times$) \\
6-layer, 1 wide & $[768,512,256,256,256,256]$ & 877k & $+11$ $[+9, +13]$ & $3{,}670$ ($1.54\times$) \\
5-layer, 2 wide & $[768,768,256,256,256]$ & 1{,}074k & $+16$ $[+15, +18]$ & $3{,}080$ ($1.83\times$) \\
\addlinespace
\multicolumn{5}{@{}l}{\emph{Cubeless money (prob5)}} \\
1-layer  & $[512]$                     & 103k & $-83$ $[-84, -82]$    & $22{,}700$ ($0.23\times$) \\
2-layer  & $[512,512]$                 & 366k & $-12$ $[-13, -11]$    & $7{,}560$ ($0.68\times$) \\
3-layer  & $[512,512,256]$             & 496k & $-4$ $[-5, -3]$       & $5{,}790$ ($0.89\times$) \\
4-layer (production) & $[512,512,256,256]$ & 562k & ---                     & $5{,}120$ \\
5-layer  & $[512,512,256,256,256]$     & 628k & $+2$ $[+1, +3]$       & $4{,}600$ ($1.11\times$) \\
5-layer, 1 wide & $[768,512,256,256,256]$ & 809k & $+5$ $[+4, +6]$       & $3{,}620$ ($1.41\times$) \\
6-layer, 1 wide & $[768,512,256,256,256,256]$ & 875k & $+7$ $[+6, +8]$   & $3{,}360$ ($1.53\times$) \\
5-layer, 2 wide & $[768,768,256,256,256]$ & 1{,}072k & $+10$ $[+9, +11]$ & $2{,}810$ ($1.82\times$) \\
\bottomrule
\end{tabular}}}
\end{table}

\pgfplotsset{
  discard if not/.style 2 args={
    x filter/.append code={
      \edef\tempa{\thisrow{#1}}\edef\tempb{#2}%
      \ifx\tempa\tempb\else\def\pgfmathresult{inf}\fi
    }
  }
}
\begin{figure}[!htb]
\centering
\begin{minipage}{0.49\linewidth}
\centering
\begin{tikzpicture}
\begin{axis}[
  width=\linewidth, height=6.2cm,
  title={\small Cubeful money},
  xlabel={Checker decisions/s on Zen 5 (log scale)},
  xmode=log, log ticks with fixed point,
  ylabel={Equity vs.\ production (mEq/game)},
  xmin=2100, xmax=36000, ymin=-220, ymax=22,
  xtick={3000,4000,6000,8000,12000,16000,24000,32000},
  xticklabels={3k,4k,6k,8k,12k,16k,24k,32k},
  y coord trafo/.code={\pgfmathparse{sign(#1)*ln(1+abs(#1))}},
  y coord inv trafo/.code={\pgfmathparse{sign(#1)*(exp(abs(#1))-1)}},
  ytick={-150,-50,-25,-10,-5,0,5,10},
  yticklabels={$-150$,$-50$,$-25$,$-10$,$-5$,$0$,$5$,$10$},
  grid=major,
  grid style={gray!20, line width=0.3pt},
  axis line style={gray!60},
  tick style={gray!60},
  tick label style={font=\footnotesize},
  label style={font=\small},
  title style={yshift=-4pt},
  axis on top,
  every node near coord/.style={font=\tiny, anchor=\anc, inner sep=3pt, text=gray!30!black},
]
\draw[gray!55!black, dashed, line width=0.6pt] (axis cs:2100,0) -- (axis cs:36000,0);
\addplot[
  color=curveblue, only marks, mark=*, mark size=1.8pt,
  mark options={fill=curveblue, draw=curveblue},
  error bars/y dir=both, error bars/y explicit,
  error bars/error bar style={line width=0.5pt, color=curveblue},
  error bars/error mark options={rotate=90, mark size=1.6pt, line width=0.5pt},
  nodes near coords=\lab, visualization depends on={value \thisrow{label}\as\lab},
  visualization depends on={value \thisrow{anc2}\as\anc},
] table[x=zen5, y=edge,
  y error plus expr={sign(\thisrow{edge}+\thisrow{ehi})*ln(1+abs(\thisrow{edge}+\thisrow{ehi})) - sign(\thisrow{edge})*ln(1+abs(\thisrow{edge}))},
  y error minus expr={sign(\thisrow{edge})*ln(1+abs(\thisrow{edge})) - sign(\thisrow{edge}-\thisrow{elo})*ln(1+abs(\thisrow{edge}-\thisrow{elo}))}]
  {capacity_cubeful.dat};
\end{axis}
\end{tikzpicture}
\end{minipage}\hfill
\begin{minipage}{0.49\linewidth}
\centering
\begin{tikzpicture}
\begin{axis}[
  width=\linewidth, height=6.2cm,
  title={\small Cubeless money (prob5)},
  xlabel={Checker decisions/s on Zen 5 (log scale)},
  xmode=log, log ticks with fixed point,
  xmin=2100, xmax=36000, ymin=-120, ymax=12,
  xtick={3000,4000,6000,8000,12000,16000,24000,32000},
  xticklabels={3k,4k,6k,8k,12k,16k,24k,32k},
  y coord trafo/.code={\pgfmathparse{sign(#1)*ln(1+abs(#1))}},
  y coord inv trafo/.code={\pgfmathparse{sign(#1)*(exp(abs(#1))-1)}},
  ytick={-80,-40,-20,-10,-5,-2,0,2,5,10},
  yticklabels={$-80$,$-40$,$-20$,$-10$,$-5$,$-2$,$0$,$2$,$5$,$10$},
  grid=major,
  grid style={gray!20, line width=0.3pt},
  axis line style={gray!60},
  tick style={gray!60},
  tick label style={font=\footnotesize},
  label style={font=\small},
  title style={yshift=-4pt},
  axis on top,
  every node near coord/.style={font=\tiny, anchor=\anc, inner sep=3pt, text=gray!30!black},
]
\draw[gray!55!black, dashed, line width=0.6pt] (axis cs:2100,0) -- (axis cs:36000,0);
\addplot[
  color=curveblue, only marks, mark=*, mark size=1.8pt,
  mark options={fill=curveblue, draw=curveblue},
  error bars/y dir=both, error bars/y explicit,
  error bars/error bar style={line width=0.5pt, color=curveblue},
  error bars/error mark options={rotate=90, mark size=1.6pt, line width=0.5pt},
  nodes near coords=\lab, visualization depends on={value \thisrow{label}\as\lab},
  visualization depends on={value \thisrow{anc2}\as\anc},
] table[x=zen5, y=edge,
  y error plus expr={sign(\thisrow{edge}+\thisrow{ehi})*ln(1+abs(\thisrow{edge}+\thisrow{ehi})) - sign(\thisrow{edge})*ln(1+abs(\thisrow{edge}))},
  y error minus expr={sign(\thisrow{edge})*ln(1+abs(\thisrow{edge})) - sign(\thisrow{edge}-\thisrow{elo})*ln(1+abs(\thisrow{edge}-\thisrow{elo}))}]
  {capacity_prob5.dat};
\end{axis}
\end{tikzpicture}
\end{minipage}
\caption{\textbf{Cost versus strength across network sizes.}  The rows of
Table~\ref{tab:capacity}, plotted as head-to-head equity at 0-ply against the
production model of each family (dashed line) versus single-thread checker
throughput on the Zen~5 machine, so that faster models lie to the right.
Labels give the number of hidden layers, with ``w'' marking a first hidden
layer widened to $768$ units and ``ww'' the first two.  The vertical axis uses a signed logarithmic
scale, $\operatorname{sign}(y) \ln(1 + |y|)$.}
\label{fig:capacity-cost}
\end{figure}

\begin{figure}[!htb]
\centering
\includegraphics[width=0.58\textwidth]{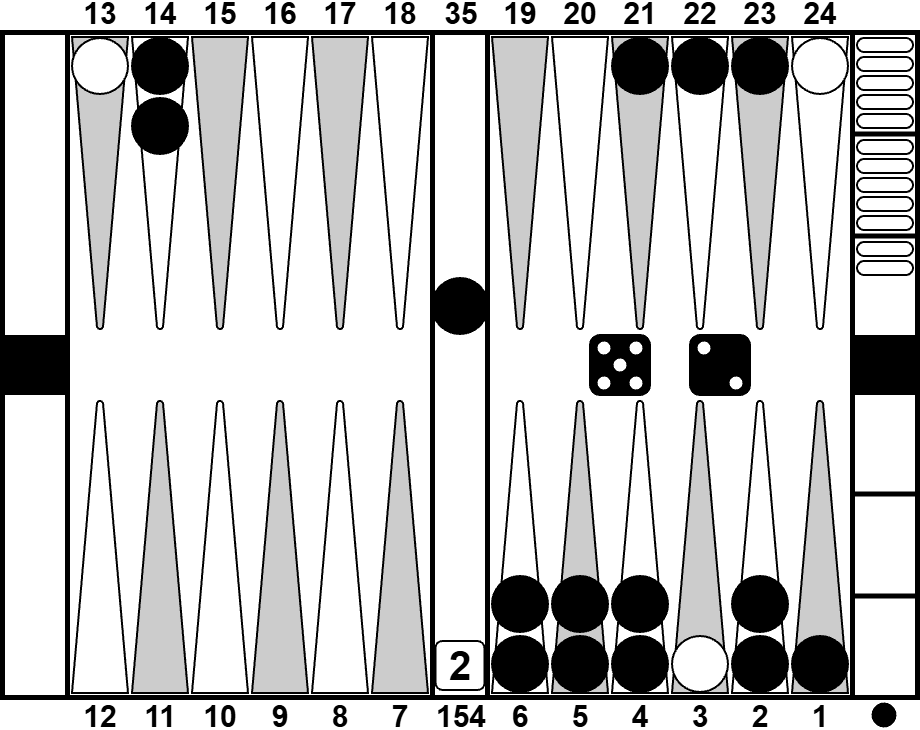}

\smallskip
{\scriptsize\nolinkurl{XGID=-ABaBBB------aB------AAAaA:1:1:1:52:0:0:3:0:10}\par}
\caption{\textbf{A creative play.}  Black to play 5-2.  At 2- and 3-ply, both our
larger cubeful and prob5 models, with hidden-layer sizes
$[768,768,256,256,256]$, slightly prefer \texttt{bar/20 5/3*}, a
``banana-split'' play.  XG++ and XG rollouts prefer \texttt{bar/23 14/9}
and judge the hit to be a large blunder, losing at least $0.2$ equity.
A \gnubg{} rollout also prefers a non-hitting play, but by a small margin.
Rollouts with our four-layer cubeful model playing at 3-ply strongly favor the
hitting play, by about $0.1$ equity.}
\label{fig:banana-split}
\end{figure}

\FloatBarrier
Figure~\ref{fig:banana-split} shows a checker play on which our
larger models and XG disagree.

\FloatBarrier
\section{Future Work}
\label{sec:future-work}

There are several natural directions for future work.  First, while the focus here has been on performance and capacity
of the neural networks at 0-ply, a systematic study
of the models with search is an important follow-up.  Second, we
restrict attention to money games (cubeless and cubeful money) and do not evaluate on
\emph{match play}, where score-aware cube strategy and the much longer horizon to
a terminal state pose additional challenges for an RL approach.  Preliminary
experiments applying PureTD to match play are promising and in progress.  Third, we do
not yet explore the kind of deep MCTS planning used by
AlphaZero~\citep{silver2018alphazero} and Stochastic
MuZero~\citep{antonoglou2022planning}, which train an evaluator under a policy
that uses substantial look-ahead.

%%%%%%%%%%%%%%%%%%%%%%%%%%%%%%%%%%%%%%%%%%%%%%%%%%%%%%%%%%%%%%%%%%%%%%%%%%%%%%%%
\section{Conclusion}
\label{sec:conclusion}

We presented PureTD, an open-source backgammon money-game engine trained
using a pure RL formulation similar to Tesauro's original TD-Gammon.  With no
look-ahead search, it achieves stronger play than our baselines, Open Sage and
\gnubg{}, even when those baselines use a one-move (1-ply) look-ahead search.  The pure RL
approach to cubeful play also proved superior to the
standard approach of approximating
cubeful equity from a cubeless model with Janowski-style heuristics, when
compared at 0-ply with equally sized networks.
Relative to prior work, our networks are larger, are trained
in batches rather than online, and learn cube action directly.

\section*{Broader Impact Statement}
We release the training pipeline and models to support reproducible
reinforcement-learning research and backgammon analysis.  Strong backgammon
agents can also enable automated wagering or undisclosed assistance,
potentially undermining fair play and causing financial harm to opponents.
Automated play and engine assistance should be disclosed to other
participants and used only where permitted by the rules of the platform or
event.

%%%%%%%%%%%%%%%%%%%%%%%%%%%%%%%%%%%%%%%%%%%%%%%%%%%%%%%%%%%%%%%%%%%%%%%%%%%%%%%%
\section*{Acknowledgments}

I am grateful to Øystein Schønning-Johansen for extensive discussions,
proofreading of the paper, and help identifying bugs in early versions of
the code, to Frank Berger for valuable feedback on the project and paper drafts,
to Todd Doucet for proofreading the paper and many insightful comments,
and to Carsten Wenderdel
(author of \texttt{wildbg}) for discussions that shaped the neural network
architecture and training settings used here.
Jacob Hilton's OCaml
implementation clarified the 1-ply amplified-equity idea, and \gnubg{}, from
the GNU Backgammon developers, served as the evaluation
baseline throughout this work.  Board diagrams were rendered with Simon
Woodhead's \texttt{bgLog NT}~\citep{woodhead2024bglog}.

\section*{LLM Usage}

Nearly all of the code for this work was written by Claude (Anthropic) using
Claude Code, under the author's direction.  The author developed the methods
and designed the experiments.
Claude also designed the optimized evaluation kernel of
Appendix~\ref{app:optimized} and assisted with editing the manuscript.

%%%%%%%%%%%%%%%%%%%%%%%%%%%%%%%%%%%%%%%%%%%%%%%%%%%%%%%%%%%%%%%%%%%%%%%%%%%%%%%%
\bibliographystyle{tmlr}
\bibliography{refs}

%%%%%%%%%%%%%%%%%%%%%%%%%%%%%%%%%%%%%%%%%%%%%%%%%%%%%%%%%%%%%%%%%%%%%%%%%%%%%%%%
\appendix
\raggedbottom
\section{Reproducibility Details}
\label{app:repro}

This appendix records the settings needed to reproduce our results.  Scripts
and exact commands are in the code repository.

\paragraph{Engine versions.}
The specific versions we used were Open Sage v2.0.20260907 (commit
\texttt{4b10c69318c3b80e2a7618232540b10a76217d20}, production model
\texttt{stage11}) and GNU Backgammon 1.08.003
(network weights MD5 \texttt{14184acc9c60ef67be0fad88548fd51a}).  The exact
commands used for all head-to-head experiments are recorded in our repository,
in \texttt{experiments/head\_to\_head}.

\paragraph{\gnubg{} configuration.}
Head-to-head games drive the unmodified \gnubg{} 1.08.003 binary through its
external-player interface.  For an $N$-ply configuration, checker and cube
evaluation are set to the same depth $N$.  All other evaluation parameters are
\gnubg's shipped defaults: noise $0.0$ (deterministic evaluation), the two-sided
bearoff database (\texttt{gnubg\_ts0.bd}) loaded, and the default
evaluation cache of $2^{19}$ entries.  \gnubg{} uses its pruning networks with
the default ``Normal'' move filter (up to eight moves within $0.16$ equity of the
best).  The unpruned comparison under ``1-ply move pruning'' below instead
disables the pruning networks and uses a keep-all move filter.

\paragraph{Open Sage configuration.}
Head-to-head runs use Open Sage through its two official interfaces: its C analysis API (\texttt{bgsage\_mobile\_analyze}, which makes
the same decisions as \texttt{BgBotAnalyzer}) for cubeful play, and
\texttt{BgBotAnalyzer} for cubeless play.  Both use Open Sage's bearoff
database.  Its ply convention is offset from ours by one, so what we report as
$N$-ply is Open Sage's ``$(N{+}1)$P'' setting, and all other parameters are its
shipped defaults.  For cubeless 1-ply evaluation, we use the default filter,
which retains the top five moves within $0.08$ equity as ranked by the 0-ply
evaluation.  When an opponent roll has more than $20$ legal replies, it keeps
the $15$ best under PubEval.  The unpruned comparison under ``1-ply move pruning'' below sets
\texttt{SAGE\_UNPRUNED=1}, widening the root filter to $1{,}000$ moves within $10.0$ equity
and effectively expanding every legal move.  It leaves the PubEval reply filter
in place.  This setting affects cubeless
play only, since the cubeful 1-ply checker play of Open Sage's analyzer and
C API has no pruning setting: it scores every
legal move, and every opponent reply to it, with the full network.  We use
this path for both playing strength and timing.  Open Sage also has a pruned
cubeful 1-ply search, which its rollouts use for checker moves (internally with a
narrower filter of two moves within $0.03$).  We run it with Open
Sage's default move filter, as in its cubeless play: it screens every legal
move by 0-ply cubeful equity, keeps at most five within $0.08$ of the best,
and re-scores those at 1-ply.  In the re-scoring, an opponent roll with more
than $16$ legal replies keeps the $15$ best under PubEval, the same thresholds
its cube decisions use.  We time this search as a separate row (``Open Sage, pruned'' in
Tables~\ref{tab:money-cost} and~\ref{tab:bench-cubeful-full}) on the same
positions, with the cube-life index and caching as in Open Sage's rollouts.  On the
benchmark's $10{,}000$ checker positions it chose the same move as
the full-width search $98.0\%$ of the time, at an average cost of $0.22$ mEq
per decision by the full-width evaluation.  We played no games
with it.  Open Sage's cube
decisions prune the moves of the player on roll as in Sec.~\ref{sec:cost}.

\paragraph{XG++ PR computation.}
XG++ PR analyses use eXtreme Gammon's per-game statistics export, which
contains two rows per game (one per player) with that player's equity
lost and unforced-decision counts.  The script
\texttt{tools/analyze\_xg\_results.py} in our repository sums equity lost over
all rows, divides by the total number of unforced decisions, and multiplies by
500 (Sec.~\ref{sec:baselines}).  The 95\% interval comes from resampling whole
games (both players together) $100{,}000$ times and taking the 2.5th and 97.5th
percentiles.

\paragraph{XG rollouts.}
The XG rollouts of Figures~\ref{fig:disagreements}
and~\ref{fig:cube-checker} use eXtreme Gammon's default rollout settings,
$1{,}296$ games with variance reduction, 3-ply checker play and XG Roller cube
decisions.

\paragraph{1-ply move pruning.}
All 1-ply results in the main text use each baseline's developer-default
move pruning.  To check that pruning does not handicap the baselines, we
repeated each 1-ply head-to-head and XG++ PR measurement with pruning
disabled, so that every legal move is expanded (Open Sage still applies its
PubEval reply filter).  Every result changed only
slightly, and none of the three pruned-versus-unpruned head-to-head contrasts
is statistically significant (Table~\ref{tab:pruning-matrix}).

\begin{table}[h]
\centering
\caption{\textbf{Effect of the baselines' 1-ply move filter.}  Each entry is
the engine's head-to-head equity (mEq/game) against PureTD 0-ply (negative
indicates a loss to our model), with 95\% confidence intervals, over $10^7$
games per cell ($2 \times 10^6$ for the Open Sage cubeful cell and the
unpruned Open Sage cubeless cell).  Cubeful entries are capped ($\pm 128$) means with Jacoby on.  The cubeful
1-ply search of Open Sage's analyzer and C API has no pruning setting, so its single cubeful
head-to-head and XG++ PR appear in the default row.  The
last column gives each configuration's cubeful XG++ Performance Rating (lower
is better) from $1{,}000$ self-play games.}
\label{tab:pruning-matrix}
{\small
\begin{tabular}{@{}llrrr@{}}
\toprule
Engine & 1-ply move filter & Cubeless (mEq/game) & Cubeful (mEq/game) & Cubeful XG++ PR \\ \midrule
\gnubg{}  & default pruning & $-12.1$ $[-12.9, -11.2]$ & $-27.2$ $[-29.0, -25.3]$ & $1.25$ $[1.20, 1.32]$ \\
\gnubg{}  & unpruned        & $-12.3$ $[-13.2, -11.5]$ & $-27.8$ $[-29.6, -25.9]$ & $1.27$ $[1.21, 1.33]$ \\
\addlinespace
Open Sage & default pruning & $-5.9$ $[-6.8, -5.1]$ & $-23.8$ $[-28.2, -19.5]$ & $1.24$ $[1.18, 1.29]$ \\
Open Sage & unpruned        & $-6.7$ $[-8.6, -4.8]$ & --- & --- \\
\bottomrule
\end{tabular}}
\end{table}

\paragraph{Computational benchmark setup.}
All numbers are reproducible in our git repository (see
\nolinkurl{experiments/benchmarks/README.md} for command lines and the
\gnubg{} build recipe).  The machines are an AMD Ryzen~5 9600X (Zen~5, Linux
6.8.0-147) and an AMD Ryzen~9 5950X (Zen~3, Linux 6.8.0-149), both running
Ubuntu 24.04 with gcc 13.3, and every engine runs on the same pinned core.
Every benchmarked configuration (an evaluator at a given search depth and game mode) uses
the same protocol, whatever its cost.
As a robustness check, we repeated the full benchmark on a second Zen~3
machine (an AMD Ryzen~9 5900X, Linux 7.0).  Its rates were lower on every
row, by a median of $4\%$ (results in
\nolinkurl{experiments/benchmarks/results/zen3_5900x_20261113/}).  On that
machine, PureTD 0-ply's throughput is still at least $3.5\times$ that of
\gnubg{} and Open Sage at 1-ply.

The fixed-position benchmark (\texttt{fixedpool\_bench.py} for the native
engines, \texttt{fixedpool\_bench\_gnubg.py} for \gnubg{}) times each
engine's per-decision call over pools of recorded decision points
committed with the repository.  Every timed region is preceded by a $5$\,s
clock ramp.  \gnubg{} is built from 1.08.003 sources with
\texttt{-{}-disable-threads} (the stock threaded build waits on a
${\sim}10$\,ms task-queue poll per evaluation, which would otherwise swamp the
true evaluation cost).  To avoid Python call overhead in the fixed-position benchmark,
\gnubg{} is timed through a 185-line additive patch to its Python module that
calls \texttt{FindnSaveBestMoves} or \texttt{GeneralCubeDecisionE} per item in
C.  Open Sage is timed by a C++ loop that calls its decision functions
directly: for cubeful play the checker and cube functions its C API runs, and
for cubeless play its move generation and \texttt{best\_move\_index}, with its
bearoff database attached.
Open Sage's shipped interfaces do further work per call that we do not time:
the C API formats a JSON reply; for a 1-ply cube decision it also runs a
second 1-ply evaluation to report cubeless probabilities and then clears a
cache.  We therefore expect the rates we report to be upper bounds on what
those interfaces deliver.  Every engine is timed inside native code around
one full decision, with no Python call in the timed region.  Open Sage and
\gnubg{} keep their evaluation caches across positions, as in normal use,
except Open Sage's full-width 1-ply cubeful decisions, which start on cleared
caches because its C API clears them after each 1-ply cube decision.  The
clearing is not timed.  PureTD has no evaluation cache.

The cubeless rows replay a shared pool of
$8{,}000$ positions (\texttt{fixedpool\_positions.json}), and the cubeful rows
replay a shared pool of $10{,}000$ checker and $6{,}000$ cube-offer decision
points (\texttt{fixedpool\_cubeful\_positions.json}) recorded from one seeded
self-play pass of an earlier production cubeful model ($253$ games).  Every row replays the full pool, so every
engine replays identical positions under identical per-position cube
states.  Among the cube-offer
decisions, about half are redouble decisions, with the cube already owned at
$2$ or higher, matching the mix in the self-play that recorded them.  Each
reported figure is the mean of eight sweeps, with $95$\% $t$-intervals over
those eight.

PureTD 0-ply's checker and cube decision throughputs are each at least
$3.5\times$ those of any 1-ply method, as claimed in Sec.~\ref{sec:intro} (the
closest case is \gnubg{} 1-ply cubeless checker decisions on Zen~5, at $3.81\times$).

\section{Complete Benchmark Tables}
\label{app:bench-tables}

Tables~\ref{tab:bench-cubeless-full} and~\ref{tab:bench-cubeful-full} report
fixed-position throughput for both game types on both machines.  For cubeful
play the benchmark measures the two decision types separately (checker play
and cube action).

\begin{table}[H]
\centering
\caption{\textbf{Cubeless-money throughput.}  Single-thread
checker-decision throughput (decisions/s).  Each figure is the mean of eight
sweeps and bracketed entries are $95$\% $t$-intervals over those eight
sweeps.  PureTD uses the PyTorch backend, and PureTD 1-ply uses the $0.08$
root move filter (Appendix~\ref{app:optimized}).}
\label{tab:bench-cubeless-full}
{\footnotesize
\setlength{\tabcolsep}{3pt}%
\begin{tabular}{@{}llrr@{}}
\toprule
Engine & Depth & Zen 3 & Zen 5 \\ \midrule
PureTD & 0-ply & $4{,}548$ $[4{,}536, 4{,}560]$ & $5{,}094$ $[5{,}060, 5{,}127]$ \\
\gnubg{} & 0-ply & $34{,}619$ $[34{,}476, 34{,}762]$ & $42{,}728$ $[42{,}573, 42{,}884]$ \\
Open Sage & 0-ply & $16{,}933$ $[16{,}849, 17{,}016]$ & $19{,}861$ $[19{,}796, 19{,}925]$ \\
\addlinespace
PureTD & 1-ply & $77.4$ $[77.3, 77.5]$ & $80.0$ $[80.0, 80.0]$ \\
\gnubg{} & 1-ply & $1{,}004$ $[1{,}001, 1{,}008]$ & $1{,}338$ $[1{,}335, 1{,}341]$ \\
Open Sage & 1-ply & $824$ $[816, 832]$ & $1{,}087$ $[1{,}083, 1{,}091]$ \\
\addlinespace
\gnubg{} & 2-ply & $53.5$ $[53.4, 53.6]$ & $72.4$ $[72.3, 72.5]$ \\
\bottomrule
\end{tabular}}
\end{table}

\begin{table}[H]
\centering
\caption{\textbf{Cubeful-money throughput.}  Single-thread
throughput (decisions/s), with checker and cube decisions reported
separately.  Each figure is the mean of eight sweeps and bracketed
entries are $95$\% $t$-intervals over those eight sweeps.  PureTD uses the
PyTorch backend, and PureTD 1-ply uses the $0.08$ root move filter
(Appendix~\ref{app:optimized}).
$^{\dagger}$Open Sage's 1-ply row is the full-width cubeful search of its
analyzer and C API.  The pruned row times the pruned cubeful 1-ply search its rollouts
use, run with its default move filter, for checker moves only
(Appendix~\ref{app:repro}).}
\label{tab:bench-cubeful-full}
\resizebox{\textwidth}{!}{%
\begin{tabular}{@{}llrrrr@{}}
\toprule
 & & \multicolumn{2}{c}{Checker dec/s} & \multicolumn{2}{c}{Cube dec/s} \\
\cmidrule(lr){3-4}\cmidrule(lr){5-6}
Engine & Depth & Zen 3 & Zen 5 & Zen 3 & Zen 5 \\ \midrule
PureTD & 0-ply & $5{,}055$ $[5{,}045, 5{,}065]$ & $5{,}638$ $[5{,}633, 5{,}643]$ & $12{,}957$ $[12{,}874, 13{,}041]$ & $24{,}513$ $[24{,}255, 24{,}771]$ \\
\gnubg{} & 0-ply & $34{,}171$ $[33{,}896, 34{,}447]$ & $41{,}607$ $[41{,}344, 41{,}870]$ & $604{,}875$ $[597{,}551, 612{,}200]$ & $714{,}856$ $[712{,}357, 717{,}355]$ \\
Open Sage & 0-ply & $16{,}727$ $[16{,}643, 16{,}810]$ & $20{,}265$ $[20{,}186, 20{,}344]$ & $276{,}195$ $[273{,}823, 278{,}567]$ & $349{,}126$ $[347{,}644, 350{,}609]$ \\
\addlinespace
PureTD & 1-ply & $62.0$ $[61.7, 62.2]$ & $63.5$ $[63.4, 63.5]$ & $134$ $[134, 134]$ & $136$ $[136, 136]$ \\
\gnubg{} & 1-ply & $1{,}094$ $[1{,}091, 1{,}097]$ & $1{,}436$ $[1{,}428, 1{,}444]$ & $3{,}173$ $[3{,}165, 3{,}181]$ & $4{,}107$ $[4{,}083, 4{,}130]$ \\
Open Sage & 1-ply & $40.2$ $[39.7, 40.7]$ & $52.0$ $[51.5, 52.4]$ & $1{,}465$ $[1{,}451, 1{,}479]$ & $1{,}873$ $[1{,}863, 1{,}883]$ \\
Open Sage, pruned$^{\dagger}$ & 1-ply & $910$ $[904, 915]$ & $1{,}197$ $[1{,}184, 1{,}209]$ & --- & --- \\
\addlinespace
\gnubg{} & 2-ply & $56.5$ $[56.4, 56.6]$ & $75.3$ $[75.1, 75.4]$ & $188$ $[188, 188]$ & $248$ $[248, 249]$ \\
\bottomrule
\end{tabular}%
}
\end{table}

\section{An Optimized Evaluation Kernel}
\label{app:optimized}

We also provide throughput data for an initial optimized version of PureTD.
The optimized kernel was designed and written by Claude.  We instructed Claude Fable 5.1 to
design a custom CPU kernel to replace the network evaluation and to
dispatch Claude Opus 5.5 agents to implement it.  Against the PyTorch path,
the kernel is about $3$--$8\times$ faster on 0-ply checker decisions,
$8$--$11\times$ faster on 0-ply cube decisions, and $3$--$8\times$ faster at 1-ply
(Table~\ref{tab:optimized}, measured under the fixed-position benchmark of
Sec.~\ref{sec:cost} in the same sweeps as Appendix~\ref{app:bench-tables}).
The main optimizations are as follows.

\paragraph{Batch the evaluations.}
For each decision, the positions to evaluate (the candidate plays from the
current board, or the opponent's replies at 1-ply) form one batch.  The
kernel splits each batch into chunks of up to $96$ rows, and each chunk into
tiles of up to $4$ rows (AVX2) or $6$ (AVX-512).  The rows of a tile are
successors of the same position, so their input vectors are nearly identical.

\paragraph{Skip zero activations.}
The board encoding is a per-point thermometer (plus a one-hot cube state for
cubeful networks), so about $85\%$ of the input entries are zero, and
$70$--$95\%$ of the ReLU hidden activations are zero, with the first hidden
layer being least sparse.  For each tile, the kernel finds the input indices $k$ at
which every row's activation is exactly zero and skips them: all terms
$w_{jk} x_k$ would be $0$, so the weights for $k$ are never loaded or
multiplied.  Skipping zero inputs in the first layer is a long-standing
technique in backgammon engines, including \gnubg's evaluator~\citep{gnubg}, and is the core
of NNUE
evaluation~\citep{nasu2018nnue,stockfish2020nnue}.  Applying the same skip
to the ReLU outputs of later layers follows Stockfish's sparse-input affine
transform~\citep{sobczyk2021sparsity,stockfish2023sparse}.
Our kernel builds one nonzero index list per tile of rows, so the weights
for each listed input are loaded once and applied to every row of the tile.

\paragraph{Incremental first layer.}
The rows of a batch are successors of one position, so each differs from
the parent in the few inputs of the moved and hit checkers (about six inputs
per successor).  The kernel computes the first layer's linear output
$z = W_0 x + b_0$, before the ReLU, once for the parent position $x$, and
obtains each successor's as $z' = z + W_0 (x' - x)$, a sum over only the
inputs that differ.  This technique is NNUE's accumulator
update~\citep{nasu2018nnue,nnuepytorch}, which HedgeHog~\citep{hedgehog2026}
calls incremental accumulation.  \gnubg's non-SIMD evaluator has, since at
least 2006, scored each candidate move incrementally from the first
candidate's hidden layer~\citep{gnubg}.  Unlike a typical NNUE engine, which carries the
accumulator along the search path and recomputes it only occasionally, our
kernel, like HedgeHog, recomputes the parent for every batch.

\paragraph{Register-blocked micro-kernel.}
The dense arithmetic uses a register-blocked outer-product
micro-kernel~\citep{goto2008anatomy,vanzee2015blis}.  The weights are packed once at load
time into panels of $16$ output units.  For each input index, the kernel
loads a few weight vectors and applies them to every row of the tile, with
the accumulators held in registers, so one load from cache feeds $4$--$6$
rows of arithmetic.  The bias and ReLU are fused into the kernel.

\paragraph{Cache-aware computation order.}
Within a layer, each group of output columns is loaded once and applied to
all tiles of a chunk before moving on, so the weights are reused from the L2
cache.

\paragraph{1-ply root move filter.}
PureTD's 1-ply throughput rows, here and in Appendix~\ref{app:bench-tables},
use a root move filter.  It screens every move at 0-ply and expands only those
that end the game or score within $0.08$ of the best move, about $5$--$6$ of the
roughly $20$ legal moves on average.  Every reply to an expanded move is scored.
Cube decisions are unpruned.
On $10{,}000$ positions from self-play games, the filtered search chose the same
move as full-width 1-ply on over $99.9\%$ of decisions.  All PureTD 1-ply strength
results in the paper are full width.

\begin{table}[H]
\centering
\caption{\textbf{PyTorch versus the optimized kernel.}  Single-thread
decisions/s for PureTD with each backend, the mean of eight sweeps (full
intervals in \texttt{experiments/benchmarks/README.md}).  1-ply uses the
$0.08$ root move filter described above.  The \gnubg{} and Open Sage
rows repeat Tables~\ref{tab:bench-cubeless-full} and~\ref{tab:bench-cubeful-full} for
comparison.  PureTD is compiled for the host CPU and uses AVX-512 on Zen 5,
while \gnubg{} and Open Sage are built with 256-bit AVX and FMA instructions,
so the Zen 3 columns compare all engines on the same instruction set.
$^{\dagger}$Open Sage's 1-ply cubeful checker entries are those of its pruned
search (the ``Open Sage, pruned'' row of Table~\ref{tab:bench-cubeful-full}).}
\label{tab:optimized}
{\small
\setlength{\tabcolsep}{5pt}%
\begin{tabular}{@{}llrrrrrr@{}}
\toprule
 & & \multicolumn{2}{c}{Cubeless checker} & \multicolumn{2}{c}{Cubeful checker} & \multicolumn{2}{c}{Cubeful cube} \\
\cmidrule(lr){3-4}\cmidrule(lr){5-6}\cmidrule(lr){7-8}
Engine & Depth & Zen 3 & Zen 5 & Zen 3 & Zen 5 & Zen 3 & Zen 5 \\ \midrule
PureTD, PyTorch & 0-ply & $4{,}548$ & $5{,}094$ & $5{,}055$ & $5{,}638$ & $12{,}957$ & $24{,}513$ \\
PureTD, optimized & 0-ply & $15{,}496$ & $36{,}835$ & $18{,}006$ & $43{,}442$ & $111{,}245$ & $264{,}045$ \\
\gnubg{} & 0-ply & $34{,}619$ & $42{,}728$ & $34{,}171$ & $41{,}607$ & $604{,}875$ & $714{,}856$ \\
Open Sage & 0-ply & $16{,}933$ & $19{,}861$ & $16{,}727$ & $20{,}265$ & $276{,}195$ & $349{,}126$ \\
\addlinespace
PureTD, PyTorch & 1-ply & $77.4$ & $80.0$ & $62.0$ & $63.5$ & $134$ & $136$ \\
PureTD, optimized & 1-ply & $222$ & $557$ & $185$ & $475$ & $420$ & $1{,}052$ \\
\gnubg{} & 1-ply & $1{,}004$ & $1{,}338$ & $1{,}094$ & $1{,}436$ & $3{,}173$ & $4{,}107$ \\
Open Sage & 1-ply & $824$ & $1{,}087$ & $910^{\dagger}$ & $1{,}197^{\dagger}$ & $1{,}465$ & $1{,}873$ \\
\bottomrule
\end{tabular}}
\end{table}

\section{Sensitivity to the Equity Cap}
\label{app:caps}

Table~\ref{tab:caps} shows the raw mean and three equity-capped means.  For
\gnubg{} at 1-ply, a single game won by PureTD at a stake of $2^{27}$
($\approx 1.3 \times 10^{8}$) points accounts for $84\%$ of the total absolute
per-game equity across $10^7$ games and moves the uncapped mean to about
$1.3 \times 10^{4}$ mEq/game.  The results against Open Sage and the
comparison of PureTD's cubeful model against Prob5Jan are unaffected, as no
game in those sessions exceeded the cap.

\begin{table}[h]
\centering
\caption{\textbf{Head-to-head means under different equity caps.}  Values are
in mEq/game from PureTD's side (every opponent lost).  The 1-ply rows use each baseline's default
move pruning, as in Table~\ref{tab:money-results}.  ``$>128$'' counts games whose
uncapped equity exceeds the $\pm 128$ cap in absolute value.  Bold marks the cap used throughout
the paper.}
\label{tab:caps}
{\small
\setlength{\tabcolsep}{5pt}%
\begin{tabular}{@{}lr rrrr r@{}}
\toprule
Matchup & Games & $\pm 128$ & $\pm 256$ & $\pm 1024$ & Uncapped & $>128$ \\
\midrule
\gnubg{} 0-ply            & $10^7$            & $\mathbf{85.2}$ & $85.3$ & $85.6$ & $99.0$ & $13$ \\
\gnubg{} 1-ply            & $10^7$            & $\mathbf{27.2}$ & $27.2$ & $27.4$ & $1.3 \times 10^{4}$ & $30$ \\
Open Sage 0-ply           & $2 \times 10^6$   & $\mathbf{79.5}$ & $79.5$ & $79.5$ & $79.5$ & $0$ \\
Open Sage 1-ply           & $2 \times 10^6$   & $\mathbf{23.8}$ & $23.8$ & $23.8$ & $23.8$ & $0$ \\
\addlinespace
Prob5Jan                  & $10^7$            & $\mathbf{14.6}$ & $14.6$ & $14.6$ & $14.6$ & $0$ \\
\bottomrule
\end{tabular}}
\end{table}

\section{Offline Analysis on Mixed-Play Games}
\label{app:mixed-play}

The offline XG++ PR analyses of Sec.~\ref{sec:results} are computed from
self-play games (same engine playing both sides).  As a robustness check, we
also analyzed $1{,}000$ head-to-head games between our cubeful money model at
0-ply and \gnubg{} at 1-ply (unpruned), scoring each engine's decisions
separately.  Table~\ref{tab:mixed-play} compares the resulting PRs with the
matching self-play values (Tables~\ref{tab:money-results}
and~\ref{tab:pruning-matrix}).  For each engine, the confidence intervals of
the two estimates overlap.

\begin{table}[H]
\centering
\caption{\textbf{Offline analysis on mixed-play games} (95\% confidence intervals).}
\label{tab:mixed-play}
\begin{tabular}{@{}lrr@{}}
\toprule
Engine & XG++ PR (self-play) & XG++ PR (head-to-head) \\ \midrule
PureTD, 0-ply & $0.80$ $[0.76, 0.84]$ & $0.86$ $[0.81, 0.93]$ \\
\gnubg{}, 1-ply (unpruned) & $1.27$ $[1.21, 1.33]$ & $1.21$ $[1.13, 1.28]$ \\
\bottomrule
\end{tabular}
\end{table}

\section{Retraining the Released Models}
\label{app:retrain}

The released production models were trained in many stages over several
months (Sec.~\ref{sec:architecture}).  The tables below give a clean recipe
that trains models of nearly the same strength from scratch.  We executed it
end to end twice with the same settings.  In head-to-head play against our
production models over $2 \times 10^7$ games per run, the 95\% confidence
interval of equity (capped at $\pm 128$ for cubeful play) contained $0$ for
both cubeful reruns and one prob5 rerun, while the other prob5 rerun trailed
production by $1.0$ $[0.4, 1.6]$ mEq/game.  All four point estimates are
within about $1$ mEq/game of production.

Both retrained models
share a common base (Table~\ref{tab:retrain-trunk}), a DMP ladder that grows
the network to its final shape followed by a task transfer to a cubeless equity
model.  The cubeful and prob5 models fork from the same final base checkpoint
and then run independently.

Each fork follows the same template of up to five stages: a warm-started
adaptation to the new task with sampled-Bellman backups, a sampled-Bellman
anneal, an exact-Bellman constant-rate stage, an exact-Bellman anneal, and a
final exact-Bellman anneal restarted at a lower rate.  We use at most
one stage of each type per fork.  Every stage uses $1{,}000$ episodes
per training round.

Throughout, \emph{type} identifies the training target (S is the
sampled-Bellman backup and E is the exact-Bellman backup, as defined in
Sec.~\ref{sec:architecture}), \emph{episodes} counts self-play games in
millions, and the two learning-rate columns give the value at the start and
end of the stage, with ``const'' meaning it was held fixed.  In the
architecture-change column, ``wu$n$'' marks a stage using an $n$-round
(learning-rate) warmup.  The depth expansions of stages \texttt{s3}, \texttt{s5}, and \texttt{s7}
use an initialization perturbation of $10^{-4}$.

\begin{table}[H]
\centering
\caption{\textbf{Shared base: DMP, then cubeless equity.}  Stages
\texttt{s1}--\texttt{s8} train the double-match-point network through the
progressive expansion of Sec.~\ref{sec:architecture}, each stage resuming from
the previous one's output.  The \texttt{cm} stages use the DMP model to warm-start a
cubeless equity model.  The base totals ${\sim}49$M episodes.}
\label{tab:retrain-trunk}
{\footnotesize
\setlength{\tabcolsep}{4pt}%
\begin{tabular}{@{}llcrrr@{}}
\toprule
Stage & Architecture Change & Type & Ep.\ (M) & Start LR & End LR \\ \midrule
\texttt{s1} & scratch $\to [80]$ & S & $1.2$ & 1e-3 & const \\
\texttt{s2} & width $[80] \to [150]$ & S & $0.6$ & 5e-5 & const \\
\texttt{s3} & depth $[150] \to [150,150]$ & S & $0.6$ & 5e-5 & const \\
\texttt{s4} & width $[150,150] \to [256,256]$ & S & $1.2$ & 5e-5 & const \\
\texttt{s5} & depth $[256,256] \to [256,256,256]$ & S & $1.2$ & 5e-5 & const \\
\texttt{s6} & width $[256,256,256] \to [512,512,256]$ & S & $3.5$ & 5e-5 & 5e-6 \\
\texttt{s7} & depth $[512,512,256] \to [512,512,256,256]$ & S & $9.0$ & 5e-5 & 5e-6 \\
\texttt{s8} & --- & E & $1.0$ & 1e-4 & 1e-5 \\
\addlinespace
\texttt{cm\_seed} & DMP $\to$ cubeless equity head (wu20) & S & $16.0$ & 5e-5 & 5e-6 \\
\texttt{cm\_const} & --- & E & $2.5$ & 1e-5 & const \\
\texttt{cm\_descent} & --- & E & $12.0$ & 1e-5 & 9e-7 \\
\bottomrule
\end{tabular}}
\end{table}

\begin{table}[H]
\centering
\caption{\textbf{prob5 cubeless model.}  Starts from the output of
\texttt{cm\_descent} of Table~\ref{tab:retrain-trunk}, a cubeless equity model at
the shared final shape.  The initial stage copies all four of the base's hidden
layers while the five-output probability head is initialized fresh.  The fork totals ${\sim}71$M episodes, $6$M sampled-Bellman
and $65$M exact-Bellman.}
\label{tab:retrain-prob5}
{\footnotesize
\setlength{\tabcolsep}{4pt}%
\begin{tabular}{@{}llcrrr@{}}
\toprule
Stage & Architecture Change & Type & Ep.\ (M) & Start LR & End LR \\ \midrule
\texttt{p5\_seed} & cubeless equity $\to$ prob5 five-output head (wu20) & S & $6.0$ & 1e-4 & 1e-5 \\
\texttt{p5\_const} & --- & E & $8.0$ & 1.8e-4 & const \\
\texttt{p5\_descent} & --- & E & $49.0$ & 1.8e-4 & 2e-6 \\
\texttt{p5\_restart} & --- & E & $8.0$ & 9e-5 & 9e-6 \\
\bottomrule
\end{tabular}}
\end{table}

\begin{figure}[H]
\centering
\begin{subfigure}[b]{0.48\textwidth}
\centering
\begin{tikzpicture}
\begin{axis}[
  width=0.95\linewidth, height=5.4cm,
  xlabel={prob5 self-play episodes (millions)},
  ylabel={Equity vs.\ \gnubg{} 0-ply (mEq/game)},
  xmin=0, xmax=73, ymin=-19, ymax=56,
  xtick={0,20,40,60},
  ytick={-10,0,10,20,30,40,50},
  grid=major,
  grid style={gray!20, line width=0.3pt},
  axis line style={gray!60},
  tick style={gray!60},
  tick label style={font=\scriptsize},
  label style={font=\scriptsize},
  axis on top,
]
% production prob5 model, Table 2 (10^7 games): 45.8 mEq
\draw[gray!55!black, dashed, line width=0.6pt]
  (axis cs:0,45.8) -- (axis cs:73,45.8);
\node[anchor=south west, font=\tiny, text=gray!35!black]
  at (axis cs:15,47.8) {production prob5 model: $45.8$};
% leg boundaries: p5_seed | p5_const | p5_descent | p5_restart
\draw[black!65, densely dotted, line width=1pt] (axis cs:6,-19) -- (axis cs:6,56);
\draw[black!65, densely dotted, line width=1pt] (axis cs:14,-19) -- (axis cs:14,56);
\draw[black!65, densely dotted, line width=1pt] (axis cs:63,-19) -- (axis cs:63,56);
\addplot[
  color=curveblue, line width=1pt, mark=*, mark size=1.2pt,
  mark options={fill=curveblue, draw=curveblue},
  error bars/y dir=both, error bars/y explicit,
  error bars/error bar style={line width=0.5pt, color=curveblue},
  error bars/error mark options={rotate=90, mark size=1.3pt, line width=0.5pt},
] table[x=ep, y=meq, y error plus=ehi, y error minus=elo] {retrain_prob5.dat};
\end{axis}
\end{tikzpicture}
\caption{prob5 fork, against \gnubg{} 0-ply.}
\label{fig:retrain-prob5-curve}
\end{subfigure}
\hfill
\begin{subfigure}[b]{0.48\textwidth}
\centering
\begin{tikzpicture}
\begin{axis}[
  width=0.95\linewidth, height=5.4cm,
  xlabel={Cubeful self-play episodes (millions)},
  ylabel={Equity vs.\ production (mEq/game)},
  xmin=0, xmax=200, ymin=-102, ymax=14,
  xtick={0,50,100,150,200},
  ytick={-100,-80,-60,-40,-20,0},
  grid=major,
  grid style={gray!20, line width=0.3pt},
  axis line style={gray!60},
  tick style={gray!60},
  tick label style={font=\scriptsize},
  label style={font=\scriptsize},
  axis on top,
]
\draw[gray!55!black, dashed, line width=0.6pt]
  (axis cs:0,0) -- (axis cs:200,0);
\node[anchor=south west, font=\tiny, text=gray!35!black]
  at (axis cs:42,2) {production parity};
% stage boundaries: cf_adapt | cf_seed | cf_const | cf_descent | cf_restart
\draw[black!65, densely dotted, line width=1pt] (axis cs:2,-102) -- (axis cs:2,14);
\draw[black!65, densely dotted, line width=1pt] (axis cs:27,-102) -- (axis cs:27,14);
\draw[black!65, densely dotted, line width=1pt] (axis cs:39,-102) -- (axis cs:39,14);
\draw[black!65, densely dotted, line width=1pt] (axis cs:154,-102) -- (axis cs:154,14);
\addplot[
  color=curveblue, line width=1pt, mark=*, mark size=1.2pt,
  mark options={fill=curveblue, draw=curveblue},
  error bars/y dir=both, error bars/y explicit,
  error bars/error bar style={line width=0.5pt, color=curveblue},
  error bars/error mark options={rotate=90, mark size=1.3pt, line width=0.5pt},
] table[x=ep, y=meq, y error plus=ehi, y error minus=elo] {retrain_cubeful.dat};
\end{axis}
\end{tikzpicture}
\caption{Cubeful fork, against production.}
\label{fig:retrain-cubeful-curve}
\end{subfigure}
\caption{\textbf{Strength during retraining.}  In (a), head-to-head equity
against \gnubg{} at 0-ply, measured at checkpoints of the prob5 fork of
Table~\ref{tab:retrain-prob5}.  The dotted lines mark the boundaries between the
four stages, and the dashed line marks the production prob5 model.  In (b),
head-to-head equity against the production cubeful model through the cubeful
fork of Table~\ref{tab:retrain-cubeful}, where zero marks parity and the dotted
lines mark the boundaries between its five stages.  Both horizontal axes omit
the ${\sim}49$M episodes of the shared base (Table~\ref{tab:retrain-trunk}).}
\label{fig:retrain-curves}
\end{figure}

\begin{table}[H]
\centering
\caption{\textbf{Cubeful-money model.}  Starts from the output of \texttt{cm\_descent} of
Table~\ref{tab:retrain-trunk}, a cubeless equity model.  The initial stage
widens the input layer from $196$ to $200$ features, the four added features
(three cube-ownership indicators and \texttt{is\_cube\_action}) being
initialized fresh, while the hidden layers and the equity head are copied
unchanged.  The fork totals ${\sim}198$M episodes, $27$M sampled-Bellman and $171$M
exact-Bellman.}
\label{tab:retrain-cubeful}
{\footnotesize
\setlength{\tabcolsep}{4pt}%
\begin{tabular}{@{}llcrrr@{}}
\toprule
Stage & Architecture Change & Type & Ep.\ (M) & Start LR & End LR \\ \midrule
\texttt{cf\_adapt} & cubeless equity $\to$ cubeful, $196 \to 200$ inputs (wu20) & S & $2.0$ & 1e-4 & const \\
\texttt{cf\_seed} & --- & S & $25.0$ & 6e-5 & 5e-7 \\
\texttt{cf\_const} & --- & E & $12.0$ & 1.8e-5 & const \\
\texttt{cf\_descent} & --- & E & $115.0$ & 1.8e-5 & 3e-7 \\
\texttt{cf\_restart} & --- & E & $44.0$ & 1.62e-5 & 2.7e-7 \\
\bottomrule
\end{tabular}}
\end{table}

\section{Continued Training with a Formal Stopping Rule}
\label{app:convergence}

The production cubeful and prob5 models were trained until their learning
curves appeared converged.  We subsequently found room for improvement in
both, and in the Lin-style ablation of
Table~\ref{tab:cube-ablation}.  This section describes additional training
under a formal stopping rule.  We then repeated the comparisons of learned
versus Janowski-style cube handling (Table~\ref{tab:janowski}) and learned
versus Lin-style cube handling (Table~\ref{tab:cube-ablation}).  Although all
three models improved, the main conclusions of the paper remain intact.

Additional training proceeded in cycles of $100$M episodes, with the learning
rate annealed linearly over each cycle as shown in
Table~\ref{tab:continued-training-lr}.
Each cycle starts from the previous cycle's final checkpoint, with
exact-Bellman backups, a fresh Adam optimizer, and $1{,}000$ episodes per round.
The first cycle starts from the original model.

\begin{table}[h]
\centering
\caption{\textbf{Learning-rate schedules for continued training.}}
\label{tab:continued-training-lr}
\begin{tabular}{@{}lrr@{}}
\toprule
Model & Initial learning rate & Final learning rate \\
\midrule
Cubeful and Lin-style ablation & 1.62e-5 & 2.7e-7 \\
prob5 & 9e-5 & 1e-6 \\
\bottomrule
\end{tabular}
\end{table}

\paragraph{Formal stopping rule.}
After each cycle, the final model plays the model at the start of that cycle
over $10^6$ games, with both models at 0-ply, using paired dice (each dice
sequence played twice, with the models swapping sides) and luck adjustment.  The
stopping rule is met when the upper end of the $95\%$ confidence interval
for the gain is below $1$ mEq/game for the cubeful networks (including the
Lin-style ablation), or below $0.5$ mEq/game in cubeless units for prob5.

\paragraph{Luck adjustment.}
For the luck-adjusted evaluations in this appendix, we subtract an estimate
of dice luck from each game's result,
using the previously trained four-layer models to estimate the equity after
each realized roll relative to the average equity over all possible
rolls~\citep{dahl2001variance,montgomery2000variance}.  This correction has
zero expectation, so the adjusted result remains unbiased even when the
model's evaluations are imperfect.  With a strong evaluator, variance can be
reduced by an order of magnitude or more, substantially lowering the number of games needed for a
given precision~\citep{montgomery2000variance,raccoonVarianceReduction}.

The Lin-style ablation was the last model to meet the stopping rule, at
cycle~5.  We therefore trained all three models for exactly five cycles
to give them the same additional training budget.  Table~\ref{tab:convergence-gates} reports
the measured gains.

\begin{table}[H]
\centering
\caption{\textbf{Stopping-rule gates.}  Each cycle's final model against the
previous cycle's final (the original model for cycle~1), in mEq/game over
$10^6$ games.  The prob5 column is measured in
cubeless money.  Brackets give $95\%$ confidence intervals.  Bold entries mark
the first cycle to meet the stopping rule.
The last row compares the cycle-5 model with the original model over
$2 \times 10^5$ games.}
\label{tab:convergence-gates}
{\small
\setlength{\tabcolsep}{6pt}%
\begin{tabular}{@{}lrrr@{}}
\toprule
Cycle & Cubeful & prob5 & Lin-style ablation \\ \midrule
1 vs.\ original & $+2.6$ $[+2.2, +2.9]$ & $+1.3$ $[+1.1, +1.4]$ & $+2.7$ $[+2.4, +3.1]$ \\
2 vs.\ 1 & $+1.6$ $[+1.3, +1.9]$ & $+0.5$ $[+0.3, +0.6]$ & $+0.9$ $[+0.6, +1.2]$ \\
3 vs.\ 2 & $+1.2$ $[+0.9, +1.5]$ & $\mathbf{+0.3}$ $[+0.1, +0.4]$ & $+1.2$ $[+0.9, +1.6]$ \\
4 vs.\ 3 & $\mathbf{+0.6}$ $[+0.3, +0.9]$ & $+0.4$ $[+0.2, +0.5]$ & $+0.9$ $[+0.6, +1.2]$ \\
5 vs.\ 4 & $+0.5$ $[+0.2, +0.8]$ & $+0.2$ $[+0.0, +0.3]$ & $\mathbf{+0.6}$ $[+0.3, +0.8]$ \\
\addlinespace
5 vs.\ original & $+6.9$ $[+6.1, +7.7]$ & $+2.7$ $[+2.3, +3.1]$ & $+5.2$ $[+4.4, +6.1]$ \\
\bottomrule
\end{tabular}}
\end{table}

The training recipes in Appendix~\ref{app:retrain} can be extended by appending
five additional cycles of $100$M episodes each, following the procedure and
learning-rate schedules described in this appendix.
We applied this extension to both independent retraining runs from
Appendix~\ref{app:retrain}.  Against the corresponding cycle-5 models reported
above, the cubeful reruns scored $-0.1$ $[-0.5, +0.3]$ and
$+0.9$ $[+0.5, +1.3]$ cubeful mEq/game.
The prob5 reruns scored $-0.5$ $[-0.7, -0.3]$ and $-0.6$ $[-0.8, -0.4]$
cubeless mEq/game, falling slightly short by amounts comparable to one or two
late-cycle gains in Table~\ref{tab:convergence-gates}, although both satisfied
the stopping rule after the fifth cycle.
All four comparisons used $10^6$ games each with paired dice and luck
adjustment, with both sides at 0-ply.

Using these extended retraining runs, we compared our cubeful model with
Prob5Jan under a larger training budget for prob5.  In each of the two runs,
we compared the
cubeful checkpoint after cycle~3 with the prob5 checkpoint after cycle~5.  Because the prob5 recipe uses $127$M fewer episodes than the
cubeful recipe, this pairing gives prob5 $73$M more training episodes
overall: $619.8$M versus $546.8$M, including the shared base.  A cubeful
episode is also generally cheaper than a prob5 episode in our implementation,
so prob5 receives at least as much training by compute as well.  Table~\ref{tab:janowski-budgets} presents the results for
run~1, where the cubeful model wins the head-to-head comparison and has the
lower XG++ PR, with most of the PR difference in cube decisions.  We also
measured head-to-head play for run~2, where Prob5Jan scored $-15.1$
$[-16.8, -13.4]$ mEq/game against the cubeful model under the same conditions.
Table~\ref{tab:cube-ablation-c5} repeats the Lin-style ablation with the
cycle-5 checkpoints of both networks.

\begin{table}[H]
\centering
\caption{\textbf{Cube-handling comparison with a larger training budget for
prob5, run~1.}  As in Table~\ref{tab:janowski}, with the cycle-3 cubeful model as
PureTD and the cycle-5 prob5 model in Prob5Jan.  The head-to-head column uses
$10^7$ games with independent dice and no luck adjustment, and the cube and
bearoff settings of Appendix~\ref{app:janowski}.  The PR games use the same
bearoff settings as in Table~\ref{tab:janowski}.}
\label{tab:janowski-budgets}
% Head-to-head: submitted paper code, adaptive x policy, old bearoff tables
% (paper ledger, origin paper_additions f78dd35c).
% Run 1: cubeful c3 7677895c vs prob5 c5 44da01ee.
% Run 2: cubeful c3 0ff8457f vs prob5 c5 98c16095.
% Run 1 XG PR: 1000 self-play games each at 0-ply, game-bootstrap CIs.
% Prob5Jan PR: adaptive x policy, 37,785 decisions (paper ledger 4434a930).
{\small
\setlength{\tabcolsep}{4pt}%
\begin{tabular}{@{}lrrrr@{}}
\toprule
Model & XG++ PR & Checker PR & Cube PR & vs.\ PureTD, 0-ply (mEq/game) \\ \midrule
PureTD   & $0.77$ $[0.73, 0.82]$ & $0.84$ $[0.79, 0.89]$ & $0.44$ $[0.35, 0.53]$ & --- \\
Prob5Jan & $1.02$ $[0.97, 1.09]$ & $0.97$ $[0.91, 1.02]$ & $1.35$ $[1.16, 1.56]$ & $-15.9$ $[-17.6, -14.1]$ \\
\bottomrule
\end{tabular}}
\end{table}

\begin{table}[H]
\centering
\caption{\textbf{Cube-handling ablation, cycle-5 models.}  As in
Table~\ref{tab:cube-ablation}, with both networks replaced by their cycle-5
checkpoints.  The head-to-head column uses $10^7$ games.}
\label{tab:cube-ablation-c5}
{\small
\setlength{\tabcolsep}{4pt}%
\begin{tabular}{@{}lrrrr@{}}
\toprule
Model & XG++ PR & Checker PR & Cube PR & vs.\ PureTD, 0-ply (mEq/game) \\ \midrule
PureTD             & $0.72$ $[0.68, 0.76]$ & $0.78$ $[0.73, 0.82]$ & $0.45$ $[0.37, 0.53]$ & --- \\
Lin-style ablation & $0.85$ $[0.81, 0.90]$ & $0.83$ $[0.79, 0.88]$ & $0.95$ $[0.80, 1.11]$ & $-9.3$ $[-11.0, -7.6]$ \\
\bottomrule
\end{tabular}}
\end{table}

The stopping rule is a practical training criterion rather than evidence of
complete convergence.  We suspect further small gains are obtainable with
additional training and stricter stopping thresholds.

\FloatBarrier

\section{Branching Factor}
\label{app:branching}

Table~\ref{tab:branching} reports the branching factor of PureTD
self-play, each row measured over $100{,}000$ games.  The branching factor of a node
is its number of children, with the legal plays counted over the $21$ distinct
dice rolls and a roll with no legal move counted as a single forced no-move.  The rows
differ in what one step (edge) of the game tree represents.

\begin{table}[!htb]
\centering
\caption{\textbf{Branching factor of backgammon money games under PureTD
self-play.}}
\label{tab:branching}
{\small
\setlength{\tabcolsep}{6pt}%
\begin{tabular}{@{}l p{6.6cm} cc@{}}
\toprule
Play & Edge & Branching factor & 95\% CI \\
\midrule
Cubeless & The on-roll player's full turn
  & $397$ & $[396, 397]$ \\
\addlinespace
Cubeful  & The on-roll player's full turn, both the cube decision and the checker play
  & $752$ & $[751, 753]$ \\
\addlinespace
Cubeful  & The on-roll player's turn, decomposed into the cube-offer decision and the checker play
  & $312$ & $[312, 313]$ \\
\addlinespace
Cubeful  & As above, and when a cube is offered, an added node for the opponent's take/pass decision
  & $242$ & $[242, 242]$ \\
\bottomrule
\end{tabular}}
\end{table}

\section{The Cube's Effect on Checker Play}
\label{app:cube-checker}

\begin{figure}[H]
\centering
\includegraphics[width=0.32\textwidth]{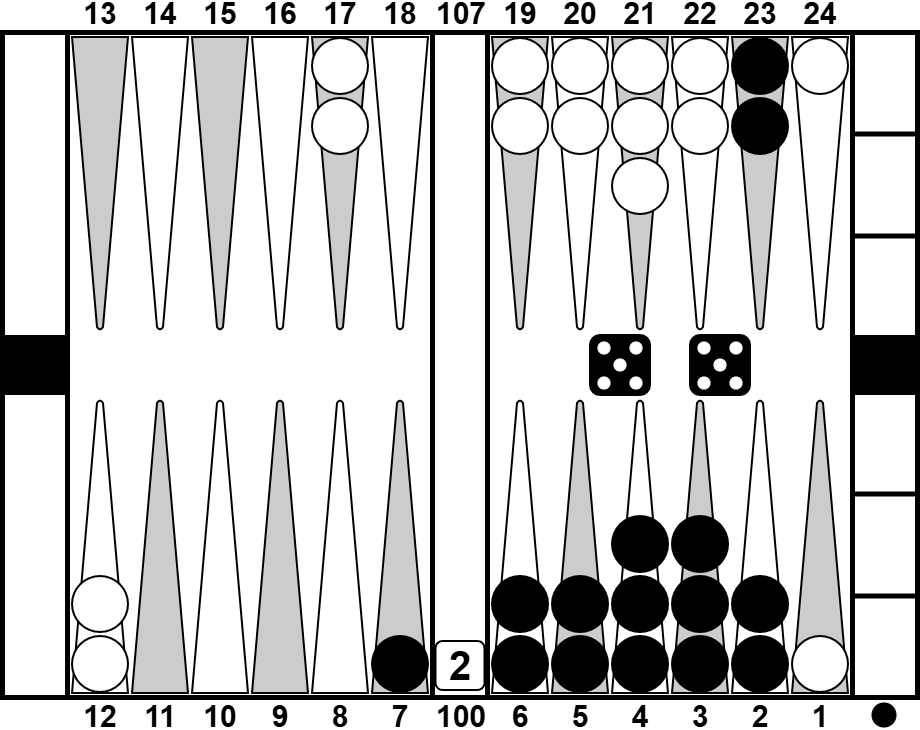}\\[3pt]
{\footnotesize
\nolinkurl{XGID=-aBCCBBA----b----b-bbcbBa-:1:1:1:55:0:0:3:0:10}\quad(money game)}\\[4pt]
{\footnotesize
\setlength{\tabcolsep}{5pt}%
\begin{tabular}{@{}ll ccc ccc cc@{}}
\toprule
& & \multicolumn{3}{c}{Win (\%)} & \multicolumn{3}{c}{Loss (\%)} &
\multicolumn{2}{c}{Equity} \\
\cmidrule(lr){3-5}\cmidrule(lr){6-8}\cmidrule(lr){9-10}
& Play & any & g & bg & any & g & bg & cubeless & cubeful \\
\midrule
\multicolumn{10}{@{}l}{\emph{Money game, cube at 2 owned by Black}} \\
1 & \texttt{23/18(2) 6/1*(2)} & 67.94 & 12.75 & 0.07 & 32.06 & 5.75 & 0.20 & $+0.428$ & $\mathbf{+0.723}$ \\
2 & \texttt{23/13(2)}         & 70.94 & 16.23 & 0.20 & 29.06 & 5.08 & 0.18 & $\mathbf{+0.531}$ & $+0.656$ \\
\addlinespace
\multicolumn{10}{@{}l}{\emph{One-point match, cube dead and gammons not counted}} \\
1 & \texttt{23/18(2) 6/1*(2)} & 67.55 & 9.93 & 0.00 & 32.45 & 4.46 & 0.13 & $+0.351$ & --- \\
2 & \texttt{23/13(2)}         & 72.05 & 13.03 & 0.42 & 27.95 & 3.51 & 0.04 & $\mathbf{+0.441}$ & --- \\
\bottomrule
\end{tabular}}
\caption{\textbf{The cube's impact on a checker-play decision.}}
\label{fig:cube-checker}
\end{figure}

The position in Figure~\ref{fig:cube-checker} arose in \emph{Blunder
Challenge}~\citep{zizka2022blunder}, a 2022 series in which Zdenek Zizka, a
two-time winner of the Ultimate Backgammon Championship, played a money game
against XG each day until he made an error larger than $0.05$ in equity.  The
two candidate plays are \texttt{23/18(2) 6/1*(2)}, which hits and attacks, and
\texttt{23/13(2)}, which runs and is the play he chose.  According to XG
analysis, the running play is better in every cubeless outcome probability: it
wins more often, wins more gammons and backgammons, and loses fewer gammons and
backgammons.  This makes it better by $0.103$ in
cubeless equity, yet it is still a $0.067$ cubeful error.  In the same position
played at DMP (one-point match), where the cube is dead and gammons do not count,
the ranking reverses and the running play is correct by $0.090$.
The table gives cubeless win and loss probabilities, including
gammons (g) and backgammons (bg), for both players.

\FloatBarrier
\section{Decision Disagreements Between PureTD and Prob5Jan}
\label{app:disagreements}

Over $20{,}000$ games of PureTD self-play (eight seeds), both PureTD and Prob5Jan
were evaluated at every decision point.  When the two disagreed, \gnubg{} at
4-ply scored the difference in equity between the two preferred plays.
Table~\ref{tab:disagreements} reports the resulting difference in mEq/game,
where a positive entry means Prob5Jan lost more equity than PureTD.  Cube actions account for more of the advantage than
checker play.

\begin{table}[h]
\centering
\caption{\textbf{Equity lost as scored by \gnubg{} 4-ply, by class of
disagreement.}  Differences are in mEq/game.}
\label{tab:disagreements}
{\small
\setlength{\tabcolsep}{8pt}%
\begin{tabular}{@{}lrl@{}}
\toprule
Decision class & $n$ & Difference per game [95\% CI] \\
\midrule
PureTD doubles, Prob5Jan does not double & $2{,}378$  & $+3.11$ $[+2.79, +3.44]$ \\
PureTD does not double, Prob5Jan doubles & $4{,}637$  & $+3.06$ $[+2.80, +3.31]$ \\
PureTD takes, Prob5Jan passes  & $152$      & $+0.04$ $[-0.06, +0.13]$ \\
PureTD passes, Prob5Jan takes  & $750$      & $-0.83$ $[-1.06, -0.61]$ \\
All cube decisions             & $7{,}917$  & $+5.38$ $[+4.90, +5.87]$ \\
\addlinespace
Checker play                   & $97{,}358$ & $+4.02$ $[+3.32, +4.69]$ \\
\midrule
Total                          &            & $+9.40$ \\
\bottomrule
\end{tabular}}
\end{table}

\FloatBarrier
\section{Janowski-Style Cube Handling in Prob5Jan}
\label{app:janowski}

Prob5Jan (Sec.~\ref{sec:cube-ablation}) combines our prob5 cubeless model with
an analytic cube layer that follows \gnubg's implementation.  For each
position, the network's five outputs are converted into an approximate cubeful
equity, and both cube decisions and checker play are derived from these
equities.

\paragraph{Notation.}
From the on-roll player's perspective, prob5 outputs the nested probabilities
$p$ (win), $p_g$ (win a gammon or better), and $p_{bg}$ (win a backgammon),
together with $l_g$ and $l_{bg}$ for the corresponding losses.  Following
\citet{janowski1993cube}, these are expressed in terms of the expected equity
conditioned on a win or a loss:
\[
W = 1 + \frac{p_g + p_{bg}}{p}, \qquad
L = 1 + \frac{l_g + l_{bg}}{1 - p},
\]
so that the cubeless (dead-cube) equity is
$E_{\mathrm{dead}} = pW - (1-p)L = p\,(W + L) - L$.  The live-cube take point
and cash point are defined as
\[
\mathrm{TP} = \frac{L - \tfrac{1}{2}}{W + L + \tfrac{1}{2}}, \qquad
\mathrm{CP} = \frac{L + 1}{W + L + \tfrac{1}{2}}.
\]

\paragraph{Cubeful equity.}
Rather than evaluating Janowski's closed-form equations directly, \gnubg{}
blends the dead-cube equity with a piecewise-linear live-cube curve,
\begin{equation}
E_{\mathrm{cubeful}} = (1 - x)\,E_{\mathrm{dead}} + x\,E_{\mathrm{live}},
\label{eq:janowski-blend}
\end{equation}
where $x \in [0, 1]$ is Janowski's cube-life index and
$E_{\mathrm{live}}$ is the piecewise-linear function of $p$ that connects the
following anchor points with straight segments:

\begin{center}
{\small
\begin{tabular}{@{}ll@{}}
\toprule
Cube state & Anchors of $E_{\mathrm{live}}$ \\ \midrule
centered      & $(0, -L),\ (\mathrm{TP}, -1),\ (\mathrm{CP}, +1),\ (1, +W)$ \\
owned         & $(0, -L),\ (\mathrm{CP}, +1),\ (1, +W)$ \\
opponent owns & $(0, -L),\ (\mathrm{TP}, -1),\ (1, +W)$ \\
\bottomrule
\end{tabular}}
\end{center}

Under the Jacoby rule with a centered
cube, gammons do not count until the cube is turned, so the endpoints (as well
as $E_{\mathrm{dead}}$) use $W = L = 1$ while $\mathrm{TP}$ and $\mathrm{CP}$
retain the full values.

\paragraph{Cube-life index.}
Prob5Jan sets $x$ by position class, as \gnubg{} does at 0- and 1-ply: $0.76$
in contact, and $0.60 + 0.0025\,n_{\mathrm{pip}}$ clamped to $[0.60, 0.80]$ in
races, where $n_{\mathrm{pip}}$ is the on-roll player's pip count.  Bearoff positions without gammon
chances take their cubeful equity from the bearoff tables and cube
calculations described
below.  Bearoff positions where a gammon is still possible use the tables'
probabilities with $x = 0.60$, \gnubg's value for one-sided bearoffs.  A
single fixed $x = 0.76$ performs within confidence bounds of this policy in our head-to-head tests
(Sec.~\ref{sec:cube-ablation}).

\paragraph{Cube decisions.}
Every cube action reduces to comparing three cubeful equities from the
perspective of the on-roll player facing a cube-offer decision, all obtained
from a single forward pass:
$\mathrm{ND} = E_{\mathrm{cubeful}}$ with the current cube state (no double),
$\mathrm{DT} = 2\,E_{\mathrm{cubeful}}$ with the opponent owning the cube
(doubled and taken), and $\mathrm{DP} = +1$ (doubled and
passed), in units of the current cube value.  Both $\mathrm{ND}$ and
$\mathrm{DT}$ are computed with
Eq.~\eqref{eq:janowski-blend} from the same network outputs and differ only in
the cube state, which enters through $E_{\mathrm{live}}$ and the Jacoby
endpoint adjustment.  Given these equities, the decisions to double, take, and
pass follow Algorithm~\ref{alg:cube-td} exactly, except that \gnubg{} and
Prob5Jan break exact ties ($\mathrm{DT} = \mathrm{ND}$) differently from
Algorithm~\ref{alg:cube-td} and PureTD.

\paragraph{Checker play.}
At 0-ply, all legal successor positions are evaluated in one batched network
call, each successor's probabilities are converted with
Eq.~\eqref{eq:janowski-blend} from the opponent's on-roll perspective at the
current cube state, and the move minimizing the opponent's cubeful equity is
chosen.  Moves are therefore ranked by cubeful rather than cubeless equity,
as in the reference engines.

\paragraph{Bearoff positions.}
In the comparison of Sec.~\ref{sec:cube-ablation}, both PureTD and Prob5Jan
use the same bearoff databases once both sides have all their checkers in
their home boards.  When each side has at most $8$ checkers left, cube and
checker-play decisions come from an exact two-sided table, computed by full
expectimax with a doubling window before every roll.  When either side has
more than $8$ checkers left but neither side can be gammoned, these decisions
come from an approximate cube calculation on bearoff distributions taken from
a one-sided table, in which each side bears off to minimize its expected
number of rolls (as in \gnubg{}).  While a gammon is still possible, Prob5Jan
converts the one-sided table's outcome probabilities, which are exact for
that bearoff policy, to cubeful equity with $x = 0.60$ by
Eq.~\eqref{eq:janowski-blend}, and PureTD uses its network.

\FloatBarrier
\section{Comparison with HedgeHog}
\label{app:hedgehog}

HedgeHog~\citep{hedgehog2026} (Sec.~\ref{sec:background}) is a backgammon
engine that includes an NNUE-style~\citep{nasu2018nnue} evaluator, of which
its developer provides an open-source implementation.  We use that
implementation with HedgeHog's two strongest publicly available models, Fox v0.34 and Aureus v0.12.  HedgeHog's production
search is not open source, and our results do not reflect on it.

\paragraph{Versions.}
HedgeHog is run from its public repository
\texttt{hedgehog-public} (code under the MIT license,
\url{https://gitlab.com/eranlambooij/hedgehog-public/}), commit
\texttt{e8847f4}.  The Fox v0.34 and Aureus v0.12 networks were downloaded
from \url{https://hedgehog-bg.com/models/} in October 2026 and are used
under HedgeHog's non-commercial terms
(\url{https://hedgehog-bg.com/community}).  Full identifiers are in
Table~\ref{tab:hh-ids}.

\begin{table}[h]
\centering
\caption{\textbf{HedgeHog repository and model identifiers.}}
\label{tab:hh-ids}
{\small
\begin{tabular}{@{}ll@{}}
\toprule
Item & Identifier \\
\midrule
Repository commit & \texttt{e8847f498f7be8798c0b71314e83c8b4075f2043} \\
\midrule
Fox v0.34 file  & \texttt{fox\_v0.34\_2026-09-26.ogxf} \\
\quad MD5       & \texttt{240a6a1605d58c44decf4fc3db413559} \\
\addlinespace
Aureus v0.12 file & \texttt{aureus\_v0.12\_2026-09-26.ogxf} \\
\quad MD5       & \texttt{379f8f20dd2b0ce05f9e24f05098fb87} \\
\bottomrule
\end{tabular}}
\end{table}

\paragraph{Head-to-head harness.}
The repository evaluates positions and does not play games, so for
head-to-head play our move generator supplies the legal successors and
HedgeHog's code ranks every one of them.
At 0-ply each successor is scored with a single network evaluation.  At 1-ply
we use the repository's expectimax search, which is unpruned and is not
HedgeHog's production search.  We use it for cubeless play only, because
cubeful 1-ply search is not supported in the public repository.

HedgeHog's networks are cubeless.  For cubeful play, we use the Janowski cube
layer from the repository (\texttt{src/cube/}, Jacoby-aware, with its default
cube-life index $x = 0.68$), which makes the double, take, and pass decisions
and ranks checker plays by cubeful equity under the current cube ownership.  This cube
layer never doubles above a cube value of $64$, so the results below use a
doubling cap of $64$.  Neither side uses a bearoff database.

\paragraph{Strength.}
Table~\ref{tab:hedgehog-strength} reports head-to-head equity against our
models, which play at 0-ply throughout.  With HedgeHog also at 0-ply, PureTD
is stronger than both Fox and Aureus in both game types, although it is slower
to evaluate with the PyTorch backend (Table~\ref{tab:hedgehog-cost}).  Under
the repository's unpruned 1-ply search, Fox's estimated equity against our
0-ply model for cubeless play is positive, but its confidence interval
contains zero.  We did not run Aureus at 1-ply, because the unpruned search
with its larger network was too slow for the number of games required.

\begin{table}[H]
\centering
\caption{\textbf{Head-to-head strength against HedgeHog.}  Each entry is the
HedgeHog model's equity (mEq/game) against PureTD at 0-ply, the prob5 model
for cubeless money and the cubeful money model for cubeful money (negative
indicates a loss to PureTD).  HedgeHog is run from its public repository,
with that repository's unpruned cubeless search at 1-ply and its Janowski cube layer
($x = 0.68$) in the cubeful columns.  Cubeful entries are capped ($\pm 128$) means
with Jacoby on, as in Table~\ref{tab:money-results}, and with the cube capped
at $64$ for both sides.  The public repository has no cubeful 1-ply search,
so that entry is omitted.  Brackets give 95\% confidence intervals.}
\label{tab:hedgehog-strength}
{\small
\setlength{\tabcolsep}{5pt}%
\begin{tabular}{@{}llrrrr@{}}
\toprule
Model & Depth & Cubeless vs.\ PureTD (mEq/game) & Games & Cubeful vs.\ PureTD (mEq/game) & Games \\ \midrule
Fox v0.34    & 0-ply & $-26.6$ $[-27.5, -25.7]$ & $10^7$          & $-68.5$ $[-70.4, -66.7]$ & $10^7$ \\
Fox v0.34    & 1-ply & $+0.8$ $[-0.6, +2.1]$    & $4 \times 10^6$ & ---                      & ---             \\
Aureus v0.12 & 0-ply & $-13.0$ $[-13.9, -12.1]$ & $10^7$          & $-47.8$ $[-49.7, -46.0]$ & $10^7$ \\
\bottomrule
\end{tabular}}
\end{table}

\paragraph{Cost.}
Table~\ref{tab:hedgehog-cost} reports single-thread throughput under the
fixed-position benchmark of Sec.~\ref{sec:cost}.  HedgeHog is timed through
its own move generation and evaluation in C++.  At 0-ply the successors of a
decision are evaluated in one batch with \texttt{eval\_ensemble\_successors},
which HedgeHog ships.  We time HedgeHog at 0-ply only, because the public
1-ply search is unpruned and is not its production search.  PureTD, re-timed in the same jobs, agrees with
Tables~\ref{tab:bench-cubeless-full} and~\ref{tab:bench-cubeful-full} within
$1.6\%$.

\begin{table}[H]
\centering
\caption{\textbf{Fixed-position throughput against HedgeHog.}  Each entry is
the mean of eight sweeps over the full pools.  The PureTD and \gnubg{} rows are those of
Tables~\ref{tab:bench-cubeless-full} and~\ref{tab:bench-cubeful-full}.  Entries
other than PureTD are followed by the per-decision cost factor relative to
the PureTD 0-ply row, as in Table~\ref{tab:cubeless-cost}.}
\label{tab:hedgehog-cost}
{\small
\setlength{\tabcolsep}{4pt}%
\resizebox{\textwidth}{!}{%
\begin{tabular}{@{}lrrrrrr@{}}
\toprule
 & \multicolumn{2}{c}{Cubeless checker dec/s} & \multicolumn{2}{c}{Cubeful checker dec/s} & \multicolumn{2}{c}{Cube dec/s} \\
\cmidrule(lr){2-3}\cmidrule(lr){4-5}\cmidrule(lr){6-7}
Engine & Zen 3 & Zen 5 & Zen 3 & Zen 5 & Zen 3 & Zen 5 \\ \midrule
PureTD, 0-ply (reference) & $4{,}548$ & $5{,}094$ & $5{,}055$ & $5{,}638$ & $12{,}957$ & $24{,}513$ \\
\gnubg{}, 0-ply & $34{,}619$ ($0.13\times$) & $42{,}728$ ($0.12\times$) & $34{,}171$ ($0.15\times$) & $41{,}607$ ($0.14\times$) & $604{,}875$ ($0.02\times$) & $714{,}856$ ($0.03\times$) \\
HedgeHog Fox v0.34, 0-ply & $21{,}803$ ($0.21\times$) & $23{,}818$ ($0.21\times$) & $23{,}012$ ($0.22\times$) & $25{,}260$ ($0.22\times$) & $295{,}332$ ($0.04\times$) & $322{,}780$ ($0.08\times$) \\
HedgeHog Aureus v0.12, 0-ply & $7{,}717$ ($0.59\times$) & $14{,}580$ ($0.35\times$) & $8{,}214$ ($0.62\times$) & $14{,}676$ ($0.38\times$) & $122{,}675$ ($0.11\times$) & $207{,}682$ ($0.12\times$) \\
\addlinespace
\gnubg{}, 1-ply & $1{,}004$ ($4.5\times$) & $1{,}338$ ($3.8\times$) & $1{,}094$ ($4.6\times$) & $1{,}436$ ($3.9\times$) & $3{,}173$ ($4.1\times$) & $4{,}107$ ($6.0\times$) \\
\bottomrule
\end{tabular}}}
\end{table}

\section{Sampled versus Exact Bellman Backups}
\label{app:backup-comparison}

We compared sampled and exact Bellman backups from the prob5 recipe's
\texttt{p5\_seed} checkpoint, immediately before the
switch to exact backups (Table~\ref{tab:retrain-prob5}).  The exact arm
followed the remaining recipe stages for $65$M episodes, while the first sampled arm
stretched those schedules to $358$M episodes, $5.5$ times as many.

% Exact- vs sampled-Bellman training at matched compute. Data: make_dat.py -> sve_*.dat (copy them next to main.tex).
% Preamble additions (main.tex already defines curveblue {HTML}{2C6FB5}):
%   \definecolor{sampledorange}{HTML}{EB6834}
%   \definecolor{tunedaqua}{HTML}{1BAF7A}
\begin{figure}[!htb]
\definecolor{sampledorange}{HTML}{EB6834}
\definecolor{tunedaqua}{HTML}{1BAF7A}
\centering
\pgfplotsset{
  sve axis/.style={
    width=0.95\linewidth, height=4.7cm,
    xlabel={Training compute (millions of exact-backup episodes or equivalent)},
    ylabel={Equity vs.\ production (mEq/game)},
    grid=major, grid style={gray!20, line width=0.3pt},
    axis line style={gray!60}, tick style={gray!60},
    tick label style={font=\scriptsize}, label style={font=\scriptsize},
    axis on top,
  },
  sve curve/.style={
    color=#1, line width=1pt, mark size=1.2pt, mark options={fill=#1, draw=#1, solid},
    error bars/y dir=both, error bars/y explicit,
    error bars/error bar style={line width=0.5pt, color=#1, solid},
    error bars/error mark options={rotate=90, mark size=1.3pt, line width=0.5pt},
  },
  sve exact/.style={sve curve=curveblue, mark=*},
  sve sampled/.style={sve curve=sampledorange, mark=square*, dashed},
  sve tuned/.style={sve curve=tunedaqua, mark=triangle*, densely dashed},
  /tikz/sve label/.style={font=\tiny, text=black!75},
}
\begin{tikzpicture}
\begin{axis}[sve axis,
  width=0.72\textwidth, height=4.5cm,
  xmin=0, xmax=68, ymin=-56, ymax=4, xtick={0,20,40,60}, ytick={-50,-40,-30,-20,-10,0},
  legend columns=3,
  legend style={at={(0.5,1.08)}, anchor=south, font=\scriptsize, draw=none},
]
\draw[gray!55!black, dashed, line width=0.6pt] (axis cs:0,0) -- (axis cs:68,0);
\addplot[sve exact] table[x=ep, y=meq, y error plus=ehi, y error minus=elo] {figures/sve_curves/sve_p5_exact.dat};
\addlegendentry{Exact}
\addplot[sve sampled] table[x=ep, y=meq, y error plus=ehi, y error minus=elo] {figures/sve_curves/sve_p5_sampled.dat};
\addlegendentry{Sampled, LR 1.8e-4}
\addplot[sve tuned] table[x=ep, y=meq, y error plus=ehi, y error minus=elo] {figures/sve_curves/sve_p5_sampled_lr.dat};
\addlegendentry{Sampled, LR 1.1e-6 (best tuned)}
\end{axis}
\end{tikzpicture}
\caption{\textbf{Prob5 training curves for exact and sampled backups.}
The exact and sampled arms ran for $65$M and $358$M episodes after the shared
checkpoint, respectively.  Scores are in cubeless mEq/game against the
production prob5 model, from $200{,}000$ games with paired dice and luck
adjustment, both sides at 0-ply.  Error bars show $95\%$ CIs.
On the x-axis, sampled episode counts are divided by $5.5$, the cost ratio
used to set the sampled budget.}
\label{fig:sve-curves}
\end{figure}

\paragraph{Relative cost.}
The cost of a training episode with sampled-Bellman backups (a sampled
episode) relative to one with exact-Bellman backups (an exact episode) depends
on the hardware and the implementation.  With the training code that preceded
the SIMD implementation and was used for most of the training in this
paper, training with sampled-Bellman backups was typically $2$--$5.5\times$
faster per episode than training with exact-Bellman backups across six
machines we measured.  We
used $5.5$, the top of that range, to set the sampled budget.  A later
data collector for sampled episodes, based on the SIMD kernel of
Appendix~\ref{app:optimized}, raised the ratio to about $18$, at which $358$M
sampled episodes cost about as much as $20$M exact episodes.  Our conclusion
does not depend on the ratio, as discussed below.

\paragraph{Learning-rate checks.}
The first sampled arm used the same learning-rate schedule as the exact arm,
which starts at 1.8e-4, and fell well below the shared checkpoint, ending at $-44$
mEq/game after $358$M episodes.  To check whether that rate was simply too
high for sampled backups, we tried seven lower initial rates (6e-5, 2e-5,
6e-6, 2e-6, 1.1e-6, 6e-7, and 2e-7), each decreasing linearly to $1\%$ of its
starting value over $358$M episodes.  Poorly performing arms were stopped at
$11$M or $44$M episodes, and two finalists, 1.1e-6 and 2e-6, ran the full
budget.  We plot the better one, 1.1e-6.  The other finished about
$1$ mEq/game below it.

Figure~\ref{fig:sve-curves} shows the results.  From the shared checkpoint,
which scored $-28$ mEq/game against the production prob5 model, the exact arm
recovered production strength after $65$M episodes.  The best tuned sampled
arm reached $-26$, an improvement of only $2.0$ mEq/game over the shared checkpoint, and
lost to the exact arm by $26$ mEq/game in direct play.
The advantage of exact backups thus persisted for this checkpoint and the
schedules tested, even with $5.5$ times as many sampled episodes.  This
advantage held no matter when the exact arm is stopped.  After only
$2$M episodes the exact arm, at $-14$ mEq/game, was already ahead of every
evaluated checkpoint of the sampled arms through $358$M episodes, and the
best tuned sampled arm changed by less than $1$ mEq/game over its last
$314$M episodes.

The same comparison for the cubeful model (not plotted) showed the same pattern.  From the corresponding
checkpoint, at $-49$ mEq/game against the production cubeful model, the exact
arm recovered production strength, the sampled arm with the same schedule
fell to $-69$, and a tuned sampled arm, stopped early, stayed near $-46$.

\end{document}